\documentclass[a4paper,fleqn]{cas-dc}

\usepackage[authoryear]{natbib}
\usepackage{amsmath,amssymb,amsfonts}
\usepackage{algorithmic}
\usepackage{graphicx}
\usepackage{textcomp}
\usepackage{amsmath,bm}
\usepackage{multicol}
\usepackage{arydshln}
\usepackage{float}
\usepackage[misc]{ifsym}
\usepackage{hyperref}
\usepackage{url}
\usepackage{color,soul}

\usepackage{multirow}
\emergencystretch=\maxdimen
\begin{document}
\let\WriteBookmarks\relax
\def\floatpagepagefraction{1}
\def\textpagefraction{.001}
\shorttitle{Interpretable Fake News Detection with Topic and Deep Variational Models}
% \shorttitle{XFN: Explainable Fake News Detection with Topic and Deep Variational Models}
\shortauthors{M. Hosseini et~al.}

\title [mode = title]{Interpretable Fake News Detection with Topic and Deep Variational Models} 
% \title [mode = title]{XFN: Explainable Fake News Detection with Topic and Deep Variational Models}                      
\tnotemark[1]
% \tnotemark[1,2]

\tnotetext[1]{MH and DA were funded by DA's University of Connecticut start-up research funds.}

% \tnotetext[2]{The second title footnote which is a longer text matter
%   to fill through the whole text width and overflow into
%   another line in the footnotes area of the first page.}

\author[1]{Marjan Hosseini}[
                        % auid=000,
                        bioid=3,
                        % prefix=Prof.,
                        orcid=0000-0002-0927-6658,
                        twitter=MarjHosseini,
                        linkedin=marjhosseini,
                        ]
\cormark[1]
% \fnmark[1,3]
\ead{marjan.hosseini@uconn.edu}
\ead[URL]{https://marjanh.com/}

% \fnmark[3]

\address[1]{Department of Computer Science and Engineering, University of Connecticut, Storrs, 06269, CT, USA.}

\credit{Conceptualization, methodology, software, validation, investigation, data curation, writing -- original draft, visualization}

\cortext[cor1]{Corresponding author}

\author[2]{{Alireza} {Javadian Sabet}}[
                        % auid=000,
                        bioid=1,
                        % prefix=Eng.,
                        orcid=0000-0001-9459-2411,
                        twitter=ArjSabet,
                        linkedin=arjsabet,
                        ]

% \fnmark[1]
\ead{alj112@pitt.edu}
\ead[url]{https://sites.pitt.edu/~alj112/}

\credit{Conceptualization, methodology, software, validation, investigation, data curation, writing -- original draft, visualization}

\address[2]{Department of Informatics and Networked Systems, University of Pittsburgh, Pittsburgh, 15260, PA, USA.}

\author[1]{Suining He}[
                        % auid=000,
                        bioid=3,
                        % prefix=Prof.,
                        orcid=0000-0003-1913-6808
,
                        % twitter=???,
                        linkedin=suining-he-9046ba40,
                        ]
% \fnmark[1,3]
\ead{suining.he@uconn.edu}
\ead[URL]{https://hesuining.weebly.com/}

% \fnmark[3]

\credit{Supervision, conceptualization, methodology, validation, investigation, writing -- review \& editing}

\author[1]{Derek Aguiar}[
                        % auid=000,
                        bioid=4,
                        % prefix=Prof.,
                        orcid=0000-0001-9166-8783
,
                        % twitter=???,
                        linkedin=derekaguiar,
                        ]
% \fnmark[1,3]
\ead{derek.aguiar@uconn.edu}
\ead[URL]{https://www.derekaguiar.com/}

% \fnmark[3]

\credit{Supervision, funding acquisition, conceptualization, methodology, validation, investigation, writing -- review \& editing}

% \cortext[cor2]{Principal corresponding author}

% \fntext[fn1]{This is the first author footnote. but is common to third
%   author as well.}
% \fntext[fn2]{Another author footnote, this is a very long footnote and
%   it should be a really long footnote. But this footnote is not yet
%   sufficiently long enough to make two lines of footnote text.}

% \nonumnote{This note has no numbers. In this work we demonstrate $a_b$
%   the formation Y\_1 of a new type of polariton on the interface
%   between a cuprous oxide slab and a polystyrene micro-sphere placed
%   on the slab.
%   }

\begin{abstract}
The growing societal dependence on social media and user generated content for news and information has increased the influence of unreliable sources and fake content, which muddles public discourse and lessens trust in the media.
%With the growth of social media and User Generated Content in the recent decade, people are more exposed to the news with unreliable sources and fake content. 
Validating the credibility of such information is a difficult task that is susceptible to confirmation bias, leading to the development of algorithmic techniques to distinguish between fake and real news.
%and spreading false information could potentially lead to losses and crimes. %dca: this is a little vague and I tried to add specific examples in the first example that might cover the 'need' statement
However, most existing methods are challenging to interpret, making it difficult to establish trust in predictions, and make assumptions that are unrealistic in many real-world scenarios, e.g., the availability of audiovisual features or provenance.
In this work, we focus on fake news detection of textual content using interpretable features and methods.
%Due to the absence of multimedia and information about the author and spread patterns in many real-world sources of news, we focus on extracting relevant features only from the textual content. 
In particular, we have developed a deep probabilistic model 
that integrates a dense representation of textual news using 
a variational autoencoder and bi-directional Long Short-Term Memory (LSTM) 
networks with semantic topic-related features inferred from 
a Bayesian admixture model. % I don't think our appoach is multi-modal. Multi-modality in ML typically refers to data from 2 or more types of sensors. If we say the name of the model here, we have to spell out the acronym. % I agree, our model is not multi modal, it leverages two models. 
Extensive experimental studies with 3 real-world datasets demonstrate that our model achieves comparable performance to state-of-the-art competing models while facilitating model interpretability from the learned topics.
% Our results demonstrate that our model achieves higher mean accuracy while providing model interpretability from the learned topics. 
Finally, we have conducted model ablation studies to justify the effectiveness and accuracy of
integrating neural embeddings and topic features both quantitatively by evaluating performance and qualitatively through separability in lower dimensional embeddings.
%Furthermore, we show that our model facilitates clustering fake and real news efficiently.  model
\end{abstract}

% \begin{graphicalabstract}
% \includegraphics[width=1\linewidth]{Diagrams/Graphical_Abstract_V2.pdf}
% \end{graphicalabstract}

% \begin{highlights}

% \item Model and predict user-generated content popularity via multi-perspective approach
% \item Identify using machine learning the features contributing to the popularity of UGC 
% \item Space-time dynamics of presence and impact on social media of the top fashion brands 
% % \item Impact and role of large events in fashion: 93\% of posts target Milan, Paris, London, New York weeks
% \item Large events role in fashion: 93\% of posts target Milan, Paris, London, New York FWs
% % \item Social media about fashion events is more affected by the host city than by the brands 
% \item Social media about fashion event is more affected by the host city than by the brands

% % \item SM users' posting dynamic about fashion weeks is directly related to the events’ time
% % \item The distribution of hashtags usage frequency is extremely heavy-tailed
% % \item Multi-modal approaches are the prevalent choice for predicting SM content popularity
% % \item The UGC reflect the in-person experience of the participants in the actual event
% % \item 15\% of 14M hashtags have been used more than 10 times during big four capitals FWs
% \end{highlights}


\maketitle

% ############################################################################################################################################

\section{Introduction}

% it's not very useful to rewrite the abstract. The intro should expand and elaborate on parts of the abstract.

% \hl{this is the background of the work, some of the observations, statements, and hypothesis can be better backed by some selected citations and references; frequent review comments would be not citing related work from the conference itself} dca: i totally agree, but we're out of time unfortunately. I do think most claims are already cited though

%what is fake news, where does it happen? 
The increased availability and consumption of user generated content on the internet has provided an ideal environment for propagating fake news, or fabricated information that imitates news media. %and is typically antagonistic and seeking to benefit from or generate controversy. %dca only abbreviate if we use many times, UGC is used twice
While fake news has existed since at least the 1\textsuperscript{st} century BC~\citep{posetti2018short}, 
the prevalence of online social media (OSM) and the echo chamber effect  of content servicing algorithms have significantly increased the number of fake news domains and their associated web traffic~\citep{cinelli2021echo}.
For instance, there was a significant increase in the number of fake news web domains prior to the 2016 U.S. presidential election~\citep{chalkiadakis2021rise} 
and an estimated 41.8\% of their traffic was driven by OSM~\citep{allcott2017social}.
False information is shared by more people and 
spreads faster than true information,  
% information, 
particularly for political topics~\citep{vosoughi2018spread}.
Platforms like Twitter and Facebook are integral to this increased exposure, 
allowing users to easily share and 
promote content without being subject to 
journalistic norms and ethical standards~\citep{posetti2018short}.
This is in stark contrast with traditional professional journalism that strives for truthfulness, accuracy, objectiveness, fairness, and accountability~\citep{ethics_standards}.
OSM are also increasingly exploited as a primary news source; 
% an estimated 
$53\%-62\%$ of U.S. adults often or sometimes rely on OSM as their source of daily news~\citep{carminati2012trust}. % that were established in response to the late 19th and early 20th century war propaganda~\cite{posetti2018short}.
% Unlike traditional media such as newspapers and TV channels, in online platforms anyone can share and spread any type of information without providing the authenticity of it. 
%OSM has thus become a preferred medium for spreading fake news and misinformation.

% why is it important? because everyone is using social media and it has a detrimental effect THIS NEEDS TO BE REWRITTEN
%Although during the birth of Web 2.0, Online Social Media (OSM) arose similar to wikis, blogs, etc.~\cite{carminati2012trust}, they soon opened their path into people's daily life in many aspects. 
% However, around $70\%$ of people can successfully validate the authenticity of the news they read~\cite{perez2017automatic}.  % dca: I think this quote is contrary to the motivation of the paper, right?
The rise of fake news and OSM as a preferred news medium has been associated with detrimental effects on society~\citep{hindman2018disinformation, lee2019global}.
In the political domain, consuming fake news is associated with decreased trust towards news media and increased political polarization~\citep{guess2020fake}.
Based on the effects of media in general, 
it has been suggested that fake news can 
encourage extremism, increase cynicism,  and apathy~\citep{lazer2018science}, and deepens belief in false claims or conspiracy theories~\citep{guess2020fake}.
% losses and crimes. 
% Additionally, since OSM is being used as the source of news for most individuals, fake news now spreads at a faster pace and has a greater impact than ever before. 
One mechanism for combating the sharing of false information is to annotate news articles with the \textit{truthfulness} of the underlying claims.
Evidence suggests that fake news headlines are considered less accurate when people are warned about the potentially false or misleading nature of the content~\citep{clayton2020real}. 

% how is fake news characterized? what types of methods exist for this? what are their limitations?
% % how is fake news characterized? what types of methods exist for this? what are their limitations?
% Due to an increased reliance on social media for news and their detrimental effects, companies like Twitter and Facebook have begun classifying shared posts 

Detecting and classifying fake news is therefore a vital goal to curtail its spread and impact.
However, we need to carefully address the following two major challenges to detect and disrupt fake news early in its propagation and to establish and retain public trust.

\textbf{Challenge 1 -- Model Interpretability while Retaining Accuracy:}
Facebook, Twitter, and other OSM have begun classifying user generated content using third party fact checking organizations such as
Snopes and Politifact and automated AI systems~\citep{babaei2019analyzing}.
News articles are either classified as a binary (real or fake) or ordinal\break categorical variable (e.g., Politifact's Truth-O-Meter) based on a level of ``realness''.
Snopes and Politifact are highly accurate and explainable, but can be slow due to the human editorial resources that are required to pass judgment~\citep{politifact, snopes}.
In contrast, automatic classification of fake news using AI is highly efficient and can be accurate~\citep{pennycook2019fighting}, but typically rely solely on ``black box'' models, i.e., 
deep neural networks~\citep{dovsilovic2018explainable}.
It is often difficult to comprehend how and why these models generated a prediction, i.e. \textit{interpret} the model, which is a precursor for understanding the model and ultimately establishing user trust~\citep{rudin2019stop, gilpin2018explaining}.
\textbf{Challenge 2 -- Missing Data Modalities:} 
% \textbf{*Check if this is correct interpretation; I thought that label is missing in this paper as well*}
%Thats true for all of them except ref  qian2018neural (Neural User Response Generator ...) which is text only and actually their focus is early detection.
AI-based fake news classification systems often 
assume the availability of specific data modalities, 
including textual content, images, user profiles (sources), network traffic, or audiovisual content~\citep{khattar2019mvae, li2014spotting, qian2018neural, wang2017liar, zhang2020fakedetector}.
However, in many real-world scenarios, acquiring many of these modalities is challenging or not possible due to data scarcity, privacy concerns, or technical limitations~\citep{STIEGLITZ2018156}. 
Many news articles do not provide images~\citep{one_2021}, and % the comments section does not apply to some websites~\cite{marchionni2015online}. % I'm not sure I understand this, comments are text base so we could handle them in principle 
although user profile information might be helpful in some scenarios, it does not necessarily characterize credibility~ \citep{moens2014mining}. 
Finally, provenance in OSM is difficult to establish since 
it is common to propagate information without mentioning the original author. % dca: can we find a citation for this?

\begin{figure*}[h!, align=\centering, width=1\textwidth]
\centering 
\includegraphics[width=0.75\textwidth]{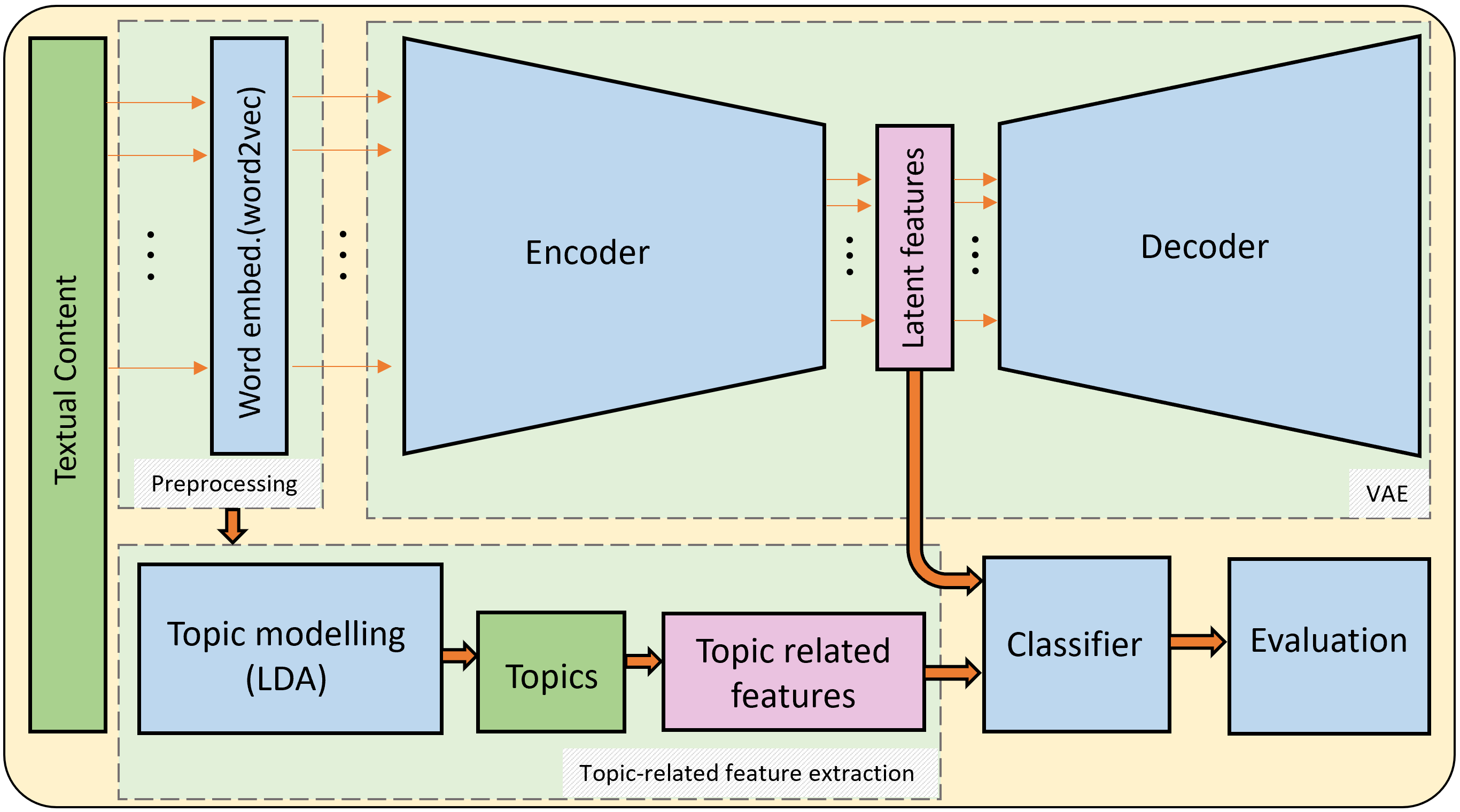}
\caption{
\textbf{Methods overview.}
Input text samples are first preprocessed (e.g., tokenized, non-English and stop words removed, and oversampled if needed). 
(Top) We use the word2vec embeddings as input to the bi-directional LSTM VAE model to extract a latent representation for each news sample.
(Bottom) After fitting the LDA model using the preprocessed text, the document specific topic distributions are combined with the latent representation from the VAE to form the input to a classifier.
}
\label{figs:model}
\end{figure*}

\subsection{Contributions}

% The contribution of this paper is two-fold. 
In this work, we consider 
% do not make strong assumptions about the availability of multiple multimedia sources; instead, we only 
automated fake news classification considering a 
data scarcity scenario 
where only textual content is available. 
We have developed a novel architecture, \texttt{LDAVAE}, that couples Bayesian topic modelling 
-- in particular, latent Dirichlet allocation (LDA) -- with a bi-directional long short-term memory (LSTM) based variational autoencoder (VAE) (Fig.~\ref{figs:model}).
%\hl{the description below should highlight some keywords for contributions and novelty}
Pretraining the bi-directional LSTM and VAE ensures our methods are accurate and can be applied quickly during test time; topic modelling provides a probabilistic mechanism for \textit{(a)} inferring the topic composition of news articles, which is often assumed to be known, \textit{(b)} using learned topics as features to improve classification accuracy, 
and \textit{(c)} interpreting the model and features. 
The learned representation is then used to classify news as real or fake. 
%Unlike many existing methods in the literature that use discrete manual subjects, LDA provides a probabilistic and automatic way of extracting the subjects.
%We integrate a deep-learning-based classifier with the VAE and formulate the problem in the supervised setting. 

% \hl{think about some key words to sell your work}
We have conducted extensive studies to compare 
\texttt{LDAVAE} with state-of-the-art (SOTA) fake news classification models using the Information Security and Object Technology (ISOT) data~\citep{ahmed2018detecting}, COVID-19 (COVID) data \citep{covid_data}, and Twitter data~\citep{boididou2018detection}.
\textit{First}, we have internally evaluated several classification methods with respect to accuracy, F1 score, false positive rates (FPR), and false negative rates (FNR) for six classifiers across each dataset. 
\textit{Second}, we have conducted an extensive ablation study to demonstrate the utility of including thematic features by comparing accuracies across each classifier for \texttt{LDAVAE}, \texttt{LDAVAE} without topic embeddings, and \texttt{LDAVAE} without VAE embeddings.
\textit{Third}, we have demonstrated highly competitive performance in terms of above-mentioned four metrics
% accuracy, F1 score, FPR and FNR 
with competing SOTA methods~\citep{khattar2019mvae, mikolov2010recurrent, wang2017liar, zhang2020fakedetector} while retaining interpretability. % you could argue the VAE features are NOT interpretable. We should probably stick to referring only to model interpretability 
%Moreover, we visualize this latent representation by dimensionality reduction techniques to show the fake and real news articles are more distinct; therefore, our framework can be helpful in unsupervised settings. 
% we show that topic modelling increases accuracy....
%clustering
%We evaluate the model on three heterogeneous datasets and provide the posterior distribution of words and topics. 
%Then provide the results of other classifiers applied to the latent representation and compare our framework with some of the similar methods. 
%Even if we use only textual data, our result is comparable with state-of-the-art (SOTA) in literature. 
% To summarize, we make the following contributions:
To summarize, we make the following four major contributions:
\begin{itemize}
    \item [\textit{i.}] designing a novel fake news detection method that combines the strengths of probabilistic and deep generative modelling (\textit{interpretability} and \textit{accuracy} as illustrated in Fig.~\ref{figs:model}) and makes few assumptions about the richness of the input: (a) only requires text and (b) infers topics instead of relying on noisy or discrete topic labels;
    % \item we make few assumptions about the richness of the input (only requires text) 
    \item [\textit{ii.}] providing a procedure for model interpretation that we demonstrate on experimental data;
    \item [\textit{iii.}] justifying the choice in classification method and using topic-based features through an extensive ablation study;
    \item [\textit{iv.}] demonstrating highly competitive performance on several metrics while providing both model and feature based interpretability. % we say highly competitive above and SOTA here. I'd rather not oversell -- table 2 is not SOTA
    % \iffalse 
    % \fi 
\end{itemize}
The source code for \texttt{LDAVAE} and evaluation scripts are freely available on GitHub\footnote{\url{https://github.com/Marjan-Hosseini/LDAVAE}}. % we should create an anonymous account and post code. If you want, we can withhold the main file and explain in the repo that we will release it if accepted.

\section{Problem Formulation and Background}
\label{sec:background}

% In this section, we provide the problem definition and relevant notation. 
% Then, we briefly explains some background concepts about the key components in this paper.

\subsection{Fake News Detection Problem: Definition and Notations}

A fake news article is a text-based document that is intentionally false or misleading~\citep{antoun2020state}. % or articles that cannot be authenticated~\cite{ajao2019sentiment}. 
In this work, fake news detection refers to the binary classification of news articles as fake or real. 
Let the training set be $\mathcal{D}_{tr} = \{d_1, \dots, d_N\}$ where each element $d_i$ is a news article (sample) indexed by $i \in \{1, \dots, N\}$. 
A news article $d_i = (x_i, y_i)$ consists of an input and class label pair where $y_i \in \{0,1\}$ and $x_i$ is a multi-dimensional vector that depends on the textual representation of the model. 
%  Our method assumes that $x_i$ is a $n_f + K$-dimensional vector where the first $n_f$ values are extracted from a VAE component and the next $K$ features are obtained from the posterior of an LDA model.  % we can't talk about this yet as we haven't introduced the VAE or LDA models
If $X=(x_1,\dots,x_i,\dots,x_N)$ and $Y=(y_1,\dots,y_i,\dots,y_N)$ are the input news articles and labels for training data, then the fake news detection problem is formally defined as learning a function $\mathcal{F}$, parameterized by $\theta$ that uses $\mathcal{D}_{tr}$ to build a classifier for predicting class labels $\hat{y}_i$, i.e., % in unseen data $\mathcal{D}_{te}$. 
\begin{align*}
    \mathcal{F}(x_i;\text{ } \theta) = \hat{y}_i,
\end{align*}

%Using these models, we can predict any future test set labels based only on the values of the features in the input data. This process requires the classifier to know the class labels in advance; that is why it is called supervised.  % dca: we don't have to explain what supervised learning is in a top tier conference paper

The objective is to minimize an arbitrary error criterion ($\epsilon$) between actual class labels and predictions (Eq.~\ref{eq:eq1}), i.e.,
% \begin{align*}
\begin{equation}
    \theta^* = \arg \min_\theta \epsilon(Y, \hat{Y}),
\label{eq:eq1}
\end{equation}
% \end{align*}
where $\hat{Y} = \{\hat{y}_1, \dots, \hat{y}_N\}$. 

% \noindent Optimal parameters in $\theta^* = \{\theta^*_{VAE}, \theta^*_{LDA} \} $  are obtained independently since $\theta^*_{VAE}$ is obtained in a supervised setting and $\theta^*_{LDA}$ in an unsupervised setting.  % this is a statement for the results section, not methods -- I moved it down
% A summary of notations used in the remainder of the paper are as follows.
% \begin{itemize}
%     \item $\mathcal{D}$: dataset, $\mathcal{D}_{tr}$: training set, $\mathcal{D}_{te}$: test set
%     \item $N$: Number of samples, indexed by $i$
%     \item $V$: The set of vocabulary detected by word2vec
%     \item $n_f$: Number of latent features obtained from encoder
%     \item $w$: word2vec dimension
%     \item $L = max\{l_i:i = 1,\dots, N\}$
%     \item $l_i$: Length of sample $i$ (number of words)
%     \item $t_i^{(j)}$: Word $j$ in sample $i$
%     % \item $\lambda_1$, $\lambda_2$: Regularization parameters
%     % \item $\lambda_1$: Regularization parameter ($=0.05$).
%     % \item $\lambda_2$: Regularization parameter ($=0.3$).
%     \item $K$: Number of topics
% \end{itemize} % we should be defining terminology where we use it, not introducing terminology before we use it, that will confuse the reader

\subsection{Latent Dirichlet Allocation}
\begin{figure}[ht]
\centering 
\includegraphics[width=0.45\textwidth]{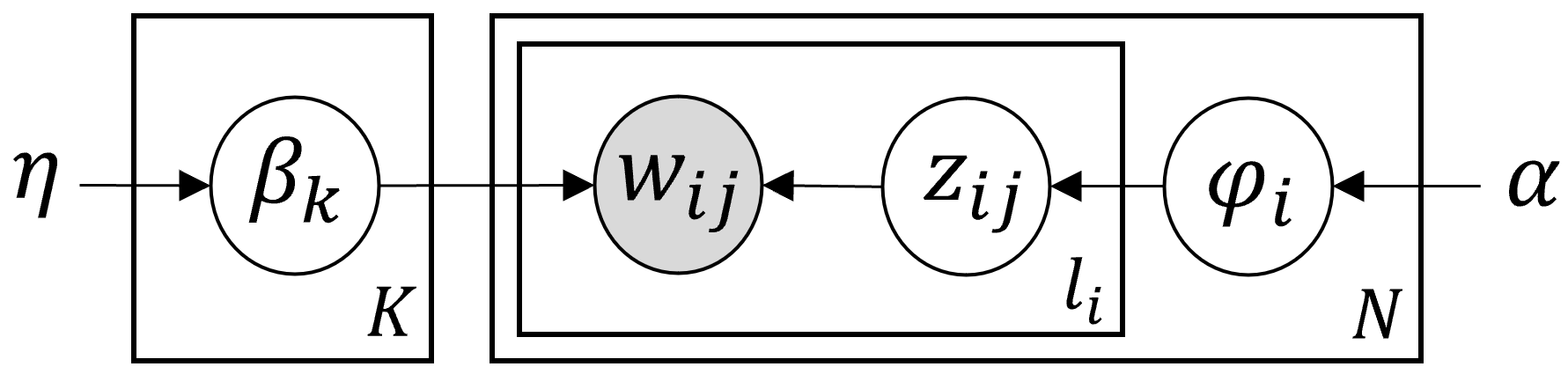}
\caption{\textbf{Probabilistic graphical model of LDA.} }
\label{figs:graphicalModel}
\end{figure}

Latent Dirichlet allocation (LDA) is an admixture model that represents topics as distributions over a word vocabulary and documents as a collection of words, each of which is sampled from a latent topic~\citep{blei2003latent}. 
Documents are then represented as distributions over topics.
%Moreover, it provides the distribution of the words in each topic. 
%These distributions can be estimated by the posterior probability of the trained parameters in the joint equation in the model. dca: we shouldn't be talking about inference yet
LDA can be defined graphically (Fig.~\ref{figs:graphicalModel}) or through its joint probability distribution (Eq.~\ref{eq:joint}).
% shows the plate model of the probabilistic graphical model with the corresponding parameters whose posterior mean values will be estimated by training the model. 
%The joint probability distribution of the model is computed in .
\begin{equation}
p(\bm{\varphi}, \bm{\beta}, \bm{Z}, \bm{W} |\eta, \alpha) = p(\bm{\beta}|\eta) p(\bm{\varphi}|\alpha)p(\bm{Z}|\bm{\varphi})p(\bm{W}|\bm{\beta},\bm{Z})
\label{eq:joint}
\end{equation}
% Table~\ref{tab:modelParam} reports the information regarding the model variables.
% First four lines of table~\ref{tab:modelParam} is the generative model, which describes the variables distributions. 
% The next four lines are the variables we know in advance.
%Random variables $\varphi$, $\beta$, $z$, and $w$ parameters whose posterior mean values will be estimated by training the model. 
Let $N$, $l_i$, and $K$ be the number of news documents, the number of words in the $i$\textsuperscript{th} news document, and the number of topics respectively.
Let matrix $\bm{W}$ be the collection of observed words $(w_{ij})$ in a news documents where $i=1,\dots,N$ and $j=1,\dots,l_i$, and $\eta$ and $\alpha$ are model hyper-parameters.
During training, we infer the posterior distributions of $\varphi_i$ (distribution of the topics in the $i^{th}$ news article), $\beta_k$ (distribution of the words in topic $k$), and $z_{ij}$ (the mapping from observed word $w_{ij}$ to a topic). 
%Matrix $\bm{Z} = (z_{ij})$ is the topic assignments for the individual words and matrix $\bm{W} = (w_{ij})$ is the observed variable (news body).

% \subsection{Term Frequency–Inverse Document Frequency (TF-IDF)} 
% It is a numerical statistic that captures the importance of words in a corpus. 
% The idea is that some words appear in all the documents frequently that are not representative of whether a document has a particular topic or not. 
% TF-IDF is proportional to a word frequency in a document; however, it is offset by the number of documents in the corpus containing that word. 

\subsection{Variational Autoencoder (VAE)} 

A VAE is an unsupervised deep neural architecture that learns to embed and reconstruct input samples~\citep{kingma2013auto}. 
It first encodes the input to the parameters of a lower dimensional multivariate normal distribution with isotropic covariance, samples from this distribution to construct a $z$ vector (i.e., the latent space), and then, the decoder reconstructs the input from $z$. 
% The difference is that in VAE, the last layer of encoder samples from the previous layer according to a normal distribution. 
The objective is derived from a variational lower bound on the marginal log-likelihood and consists of maximizing the sum of an expected log-likelihood term $p(\bm{x}|\bm{z})$ and the negative Kullback–Leibler (KL)  divergence between the posterior and prior distributions $D_{KL}(q(\bm{z}|\bm{x})||p(\bm{z}))$.

\subsection{Long Short-Term Memory} 
Long short-term memory (LSTM) is a recurrent neural network (RNN) developed to address the problem of vanishing gradients and to accommodate longer-term dependencies than traditional RNNs~\citep{hochreiter1997long}. 
%The main incentive for developing LSTM was the error flow in RNNs and addressing the problem of vanishing gradient and long time lags in methods that deal with sequential data, such as traditional RNNs and HMMs. 
Long-term dependencies with arbitrary gaps are modelled using a series of LSTM units that consist of input, output, and forget gates. 
% and addresses the modelling of long-term dependencies by remembering data with arbitrary gaps using its three gates. 
% If the sequence of state $h$ in LSTM are indexed by $t$, then we have
% \begin{flalign*}
%     &[f_t, i_t, o_t] = \sigma(Wh_{t-1} + Ux_t + b)\\
%     &c_t = f_t \odot c_{t-1} + i_t \odot (Wh_{t-1} + Ux_t + b) \\
%     &h_t = o_t \odot tanh(c_t)    
% \end{flalign*}
% where $\sigma$ is the sigmoid function, $f_t$, $i_t$, $o_t$ and $c_t$ are forget, input and output gates and memory state at $t$ and $x_t$ is the input. $W$ and $U$ are weight matrices and $b$ is the bias vector and $\odot$ denotes element-wise matrix multiplication.
A \textit{bidirectional-LSTM} combines two independent recurrent layers side-by-side, one of which receives the input sequence while the other receives a reversed copy~\citep{graves2005framewise}.
By processing both the forward and backward directions simultaneously, bidirectional-LSTMs exhibit improved performance over traditional LSTMs for sequence classification problems~\citep{liu2019bidirectional}.
%improves traditional LSTMs in terms of model performance on sequence classification problems.  This enhancement combines two independent recurrent layers .

\subsection{Word2vec} 
Standard approaches to representing words and documents consider feature vectors in a high dimensional vocabulary $V$.
The word2vec model is a two-layer neural network that learns word embeddings that attempts to maximize $\log P(w_O|w_I)$, or the log probability of a given word $w_O$ given an input word $w_I$~\citep{mikolov2013distributed}.
After training on a corpus, word2vec embeds words into a significantly lower dimensional space $w$, such that the cosine distance of semantically similar word embeddings is small and semantically dissimilar words is high.
\subsection{Dimensionality Reduction} 
Dimensionality reduction methods are designed to transform high-dimensional data into a lower-dimensional space while attempting to preserve some meaningful properties of the original data~\citep{cunningham2015linear}. 
Unlike feature selection methods that seek to subselect a set of features without augmenting them, dimensionality reduction methods transform features through this lower dimensional mapping.
They are typically applied for low dimensional visualizations and exploration of data and results, data compression or denoising, or as a step prior to unsupervised or supervised learning~\citep{rosipal2001kernel}.
In this work, we consider two dimensionality reduction methods: principle component analysis and t-distributed stochastic neighbor embedding.

%\subsubsection{Principle Component Analysis (PCA)}
Principle component analysis (PCA) is a linear dimension reduction method that uses an eigendecomposition to factorize data covariance matrix in terms of its eigenvectors and eigenvalues and find the axes in which data has more variance~\citep{ringner2008principal}. 
Then it projects the data points to these new orthogonal axes such that the greatest variance by any projection of the data comes to lie on the first coordinate. 
In other words, it transforms many correlated variables into a smaller number of uncorrelated variables (principal components).
The first principal component accounts for as much of the variability in the data as possible, and each succeeding component accounts for as much of the remaining variability as possible. 
However, it is unable to capture non-linear relationship between variables.

%\subsubsection{T-Distributed Stochastic Neighbor Embedding (t-SNE)}
T-distributed stochastic neighbor embedding (t-SNE) is a non-linear probabilistic dimensionality reduction method  designed for visualizing high dimensional data; it is based on the idea that similar objects in the high dimensional space should be represented proportionally closer to each other than dissimilar objects in lower-dimensional space~\citep{van2008visualizing}. 
To accomplish this, t-SNE minimizes the KL divergence between a joint probability distribution, in the high dimensional space and a joint probability distribution in the low dimensional space. 
The t-SNE cost function is a symmetrized version of the SNE cost function with simpler gradients and a Student t distribution to compute the similarity between pairs of points, which is more appropriate for data with outliers. 
Unlike PCA, t-SNE captures the non-linear relationships between random variables focusing on local structure, but does not preserve global distances or densities~\citep{narayan2021assessing}. %, with the cost of more computational complexity. 

% \subsection{Synthetic Minority Over-sampling Technique (SMOTE) for categorical data}
% Not sure if we needed to write about it

% other techniques:
% https://www.sciencedirect.com/science/article/pii/S1877050919314152

% \break

\section{Prior Work in Fake News Detection}
\label{sec:SOTA}
% 1. https://ieeexplore.ieee.org/stamp/stamp.jsp?tp=&arnumber=9101457

% 2. MVAE: https://dl.acm.org/doi/10.1145/3308558.3313552

Methods for classifying fake news employ a subset of three primary strategies: propagation-based, source-based and content-based.
The motivation for propagation-based techniques is that the diffusion and spread of fake news through social media differs systematically from real news in terms of the speed and the patterns of propagation~\citep{vosoughi2018spread, jin2014news}. 
Source-based methods use features derived from the provenance of news articles, including the general behavior of the source in OSM~\citep{baly2018predicting}. 
When available, detection using source-based methods can be accurate and fast, however it is often the case that information concerning the spread of news or its authorship is not observed for other users; even when observed, source-based methods can suffer from legitimate users unintentionally spreading fake news that was initiated from other sources~\citep{10.1145/3395046}. % this should be cited
%Considering the type of information the fake news detection algorithms in the literature require, these methods  fall into three main categories, propagation-based, source-based and content-based, or a combination of them.
We elaborate more thoroughly on \textit{content-based approaches} since this is the primary focus of \texttt{LDAVAE}.
%We bring a brief overview of each category, but since the proposal of this work fits the content-based category, we provide more information on the related work which followed the same strategy and their proposed models can be adjusted
% modified such that enable us 
%for comparison under the same situation.
%Also, to familiarize readers with other strategies, we briefly mention some of the state-of-the-arts (SOTA) in fake news detection that consider different features.

% Fake news detection techniques can be generally categorized into three types of approaches according to which type of information is extracted for detecting if the news is fake or not. 
% Propagation-based, source-based
% and content-based. 
% In the following ....

% ==============================
% It is not worth a subsection if it's just 1 or 2 sentences.
%\subsection{Propagation-based}

% ==============================
%\subsection{Source-based}

% Source-based methods benefit the information about the source of the news. 
% This information includes the general behavior of the source in social media and is able to detect false news fast and efficient~\cite{baly2018predicting}. 
% However, in many cases, the information about the spread of the news or the author is not provided and even if provided, it neglects that some real users can unintentionally spread some of the false news started from other sources.

% ==============================
% \subsection{Content-based Methods}

Content-based techniques extract features from the content of news such as text, images, or audiovisual content \citep{khattar2019mvae, wang2017liar, zhang2020fakedetector}. 
This approach assumes that features such as the language, topic, and style of the news body are discriminative attributes for validating its authenticity~\citep{afroz2012detecting, rashkin2017truth, rubin2016fake}. 
% On the other hand, content-based techniques focus on the characteristics of the news to extract different types of features, for example, linguistic, i.e., lexical or syntactic, characteristics. 
% This type of approach assumes that some features such as the language, topic, and style of the news body are discriminative attributes for validating its authenticity~\cite{afroz2012detecting, rashkin2017truth, rubin2016fake}. 
Here, we focus on the textual news content since this information is typically always available for news articles. 
%requires minimum available information on each news. 

Textual content-based features can be categorized into three main types. 
\textit{Syntactic features} include statistical information about the sentences, like sentence complexity and the frequency of different parts of speech or specific patterns. 
\textit{Lexical features} concern the usage of specific words or phrases in the texts, such as bi-grams and tri-grams. 

% Because of the above-mentioned reasons, in this study, we focus on the content of the news rather than the source or propagation-based techniques.
% Content-based features can be categorized into three main types. 

% \subsubsection{Syntactic features}

% \subsubsection{Lexical features}

% \subsubsection{Semantic features}
\textit{Semantic features} correspond to the meaning surrounding text and are extracted using techniques from natural language processing (NLP) and data mining such as sentiment analysis \citep{liu2010sentiment} and emotion mining \citep{yadollahi2017current}. 
% They refer to sentimental characteristics of the content and they are usually extracted using advanced Natural Language Processing (NLP) and data mining techniques such as sentiment analysis, opinion, and emotion mining approach. 
Recently, extracting word embeddings \citep{mikolov2013distributed} and topics from text~\citep{ito2015assessment} has been proposed as potentially useful features for supervised learning.
% However, employing them in fake news detection methods gain relatively less attention compared to other characteristics. %dca: i don't feel comfortable with this sentence if it's not justified.
% In this regard, we would like to include these types of features in our method. 
Features can be extracted manually through domain knowledge, but this task is tedious and subject to bias; in contrast, deep learning methods learn features automatically from neural embeddings of the textual content, which can be done more efficiently and with less human bias~\citep{ma2016detecting}. 
%To produce efficient and compact features, DL approaches concern is designing suitable layers and architecture of the network to capture all types of contents.  % dca: i don't know what this means
Neural embeddings are then typically used as input for machine learning classifiers, such as support vector machines (SVM), random forests (RF), decision trees, or logistic regression (LR), or deep learning methods such as RNNs~\citep{mikolov2010recurrent} or convolutional neural networks (CNN)~\citep{bondielli2019survey}.

Based on these prior studies, various approaches based on deep learning components have been studied.
The Multi-modal Variational Autoencoder for Fake News Detection (MVAE) uses a shared representation of the news for further classification as fake or real based on multi-modal variational autoencoder~\citep{khattar2019mvae}.
FakeDetector uses latent representation of the text, text subject, and information about the authors profile (e.g., their title, job, and credibility) \citep{zhang2020fakedetector}. 
The Hybrid CNN is a deep learning based method, with parallel CNN and LSTM layers  for processing different modalities~\citep{wang2017liar}. %IMO it's not a good idea to use citations as elements of sentences. Styles change how they are presented. Here, the reader doesn't know the method name so we have to redefine in the results.

% I added most of the 
Our proposed work differs from most of the prior studies in the following three aspects:
\textit{(1)} Probabilistic model parameters in \texttt{LDAVAE} are interpretable as topic distributions for each document and word distributions for each topic in a news articles corpus; the posterior means of learned parameters are used as additional features for classification.
\textit{(2)} Our model requires \textit{only} textual data.
\textit{(3)} We retain excellent efficiency by integrating a deep architecture (VAE) with LDA. % dca: Which prior models are innefficient? We are saying that this separates us from some models, which ones? I really wish we had some runtime or memory usage results... a reviewer is going to kill us because we make this claim without justifying it. We could remove this claim and just go with two.

\section{Proposed Method}
\label{sec:sysdesign}

% In this section, we explain and motivate the method, describe the individual components and the overall information flow through the framework. 
%The architecture is based on a multi-modal classification of news as real or fake. 
%Even though, our method uses only the textual content of the news, we consider it as a multi-modal model, because we use two types of attributes, namely textual characteristics and subject as features. 
%We learn the subjects automatically.
Our method, \texttt{LDAVAE}, is based on two views of news article text: a deep neural embedding and a probabilistic topic embedding.
\texttt{LDAVAE} uses a VAE for extracting a lower dimensional semantic representation of the news article and an LDA model for extracting topic-based features. 
% Our motivation for considering textual content only is that visual content and other sources of information such as user profile, comments, and the spread pattern of the news is not always available or costly to that not all of the datasets include associated multimedia such as image or video or other types of attributes. % dca: we've already motivated this, can reenable with more time
% So here, we focus on the latent clues in the textual content. Besides, 
% since VAE is not originally designed to capture information such as topics, 
The LDA model differs from the VAE in that the parameter values of the latent variables are highly interpretable. 
Our motivation for combining the two representations is that they capture similar but complementary discriminative features for fake news classification that will ultimately increase model performance, while retaining interpretability of model parameters.

% We share the code publicly available in the following repository\footnote{\href{https://github.com/Marjan-Hosseini/Explainable-FakeNews-Detector}{https://github.com/Marjan-Hosseini/Explainable-FakeNews-Detector}}.

% Fig.~\ref{figs:model} is an overview of the flow of the information in our method. Each box represents either a task/component or an important input/output.

% \begin{figure}[h]
% \centering 
% \includegraphics[width=0.5\textwidth]{Figs/Model_comp.PNG}
% \caption{General overview of the model. In the upper part, the data is pre-processed, oversample the minority class if needed, transformed using the word2vec model, and enters the VAE unit, where the latent features are extracted. Meanwhile, in the lower part, the topics are obtained through training the LDA model on pre-processed data. Then the latent and topic-related features are concatenated and being used by the classifier.}
% \label{figs:model}
% \end{figure}

% \begin{figure}[]
% \centering 
% \includegraphics[trim=0 0 0 0,clip,width=0.49\textwidth]{Figs/Bilstm3.PNG}
% \caption{\textbf{Bi-LSTM layer on a fake news article from the COVID data.} The past and future information for the $j$\textsuperscript{th} word $t^{(j)}$ in the news article is collected by \textit{forward} (bottom) and \textit{backward} (top) LSTM units.}
% \label{figs:bilstm}
% \end{figure}

\subsection{Variational Autoencoder}
\begin{figure}[ht]
\centering 
\includegraphics[trim=0 0 0 0,clip,width=0.49\textwidth]{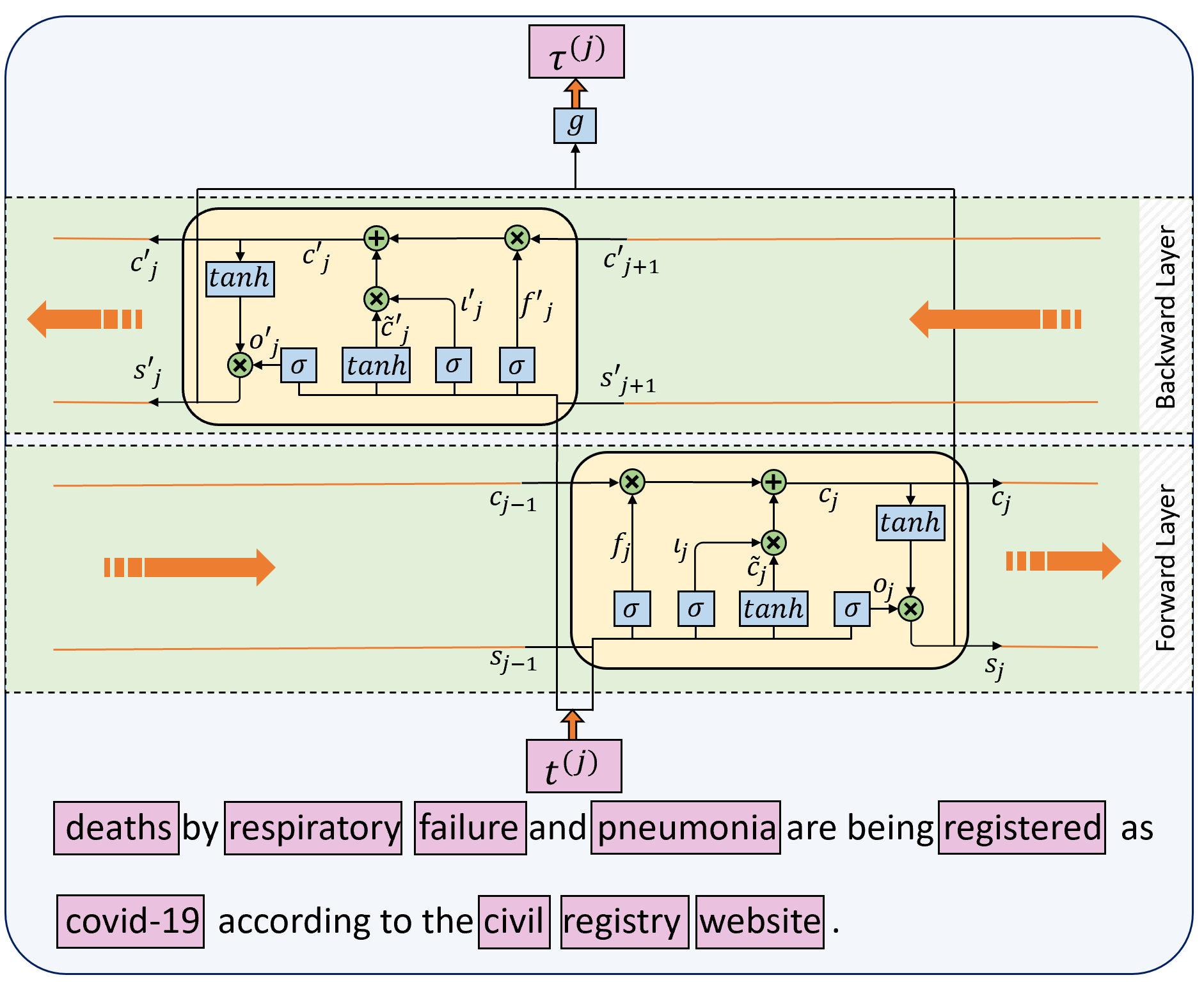}
\caption{\textbf{Bi-LSTM layer on a fake news article from the COVID data.} The past and future information for the $j$\textsuperscript{th} word $t^{(j)}$ in the news article is collected by \textit{forward} (bottom) and \textit{backward} (top) LSTM units.}
\label{figs:bilstm}
\end{figure}

The VAE component in our model, composed of an encoder and decoder, is extended to include a classifier and corresponding cross-entropy loss~\citep{khattar2019mvae}.
% We add a classifier component to regular VAE to make it supervised. 
%This architecture has been used before~\cite{khattar2019mvae}, one of the main components in our model, is VAE, which is composed of an encoder, a decoder, and a classifier. 
The architecture of the encoder and decoder is composed of bi-directional LSTM (Bi-LSTM; Fig.~\ref{figs:bilstm}) and fully connected layers. 
% Our motivation for using Bi-LSTM is that 
% sequential layers often suffer from a problem of vanishing gradients which might lead to learning inefficient dependencies between the words by the model. LSTM layers, however, overcome this problem by including the forget gate and hidden states. 
% Furthermore, 
%Bi-LSTM improves the accuracy over the traditional LSTM.  % dca: we already say this, and this claim is uncited here
% The operations we apply in the Bi-directional scheme is depicted in Fig. \ref{figs:bilstm} and Eq. \ref{eq:lstm1} and \ref{eq:lstm2}.
The Bi-LSTM architecture introduces a new layer of LSTM units that processes the sequence tokens in the backward direction. 
This overcomes the limitations of a traditional scheme by preserving the information from the past in the forward layer and from the future in the backward layer.
% In our model, the $j$\textsuperscript{th} word $t^{(j)}$ in the news article , this information are concatenated by the forward and backward LSTM units. % I don't think this sentence is doing anything and it was hard to read
In the forward unit of our model (Fig.~\ref{figs:bilstm} bottom), the information from words positioning before the $j$\textsuperscript{th} word token $t^{(j)}$ are passed through $s_{j-1}$ and $c_{j-1}$, i.e.,
\begin{flalign}
    &[f_j, \iota_j, o_j] = \sigma(\mathcal{W}s_{j-1} + Ut^{(j)} + b),\nonumber\\ 
    & \tilde{c}_j = \text{tanh}(\mathcal{W}s_{j-1} + Ut^{(j)} + b),\nonumber\\
    & c_j = f_j \odot c_{j-1} + \iota_j \odot \tilde{c}_j,\nonumber\\
    & s_j = o_j \odot \text{tanh}(c_j).    \label{eq:lstm1}
\end{flalign}

For the same word $t^{(j)}$ future words are taken into consideration in the backward LSTM unit (upper part in Fig.~\ref{figs:bilstm}). The information from the future are passed through $s'_{j+1}$ and $c'_{j+1}$, i.e.,
\begin{flalign}
    &[f'_j, \iota'_j, o'_j] = \sigma(\mathcal{W}'s'_{j+1} + U't^{(j)} + b'),\nonumber\\
    & \tilde{c}'_j = \text{tanh}(\mathcal{W}'s'_{j+1} + U't^{(j)} + b'),\nonumber\\
    & c'_j = f'_j \odot c'_{j+1} + \iota'_j \odot \tilde{c}'_j, \nonumber\\
    & s'_j = o'_j \odot tanh(c'_j),    \label{eq:lstm2}
\end{flalign}
where $\sigma$ is the sigmoid function, $f_j$, $i_j$, $o_j$ and $c_j$ are forget, input and output gates and memory state at position $j$ in the input text. $\mathcal{W}$, $\mathcal{W}'$, $U$ and $U'$ are weight matrices and $b$ and $b'$ are the bias vectors and $\odot$ denotes element-wise matrix multiplication. 
Then, the output is $\tau^{(j)} = g(s_j, s'_j) $. 
Here $g$ concatenates  $s_j$ and $s'_j$ (Fig. \ref{figs:bilstm}).

% \begin{flalign}
%     & \tau^{(j)} = g(s_j, s'_j)  
% \end{flalign}
% Here $g$ concatenates  $s_j$ and $s'_j$.
%A VAE was originally defined an unsupervised model that reconstructs the input by maximizing a lower bound on the marginal log-likelihood. 
To incorporate labels, we couple a classifier along with the encoder and decoder such that during training the parameters are optimized with respect to the unsupervised VAE ($\mathcal{L}_{rec}$) and classifier ($\mathcal{L}_{BC}$) losses.
The VAE loss is given by:
\begin{align}
    \mathcal{L}_{CE} &= - \mathbb{E}_{i \sim  \mathcal{D}}\left[\sum_{j=1}^{l_i} \sum_{v \in V} 1_{v = t_i^{(j)}} \log t_i^{(j)}\right], \nonumber \\
    \mathcal{L}_{KL} &=  - \frac{1}{2} \sum_{f=1}^{n_f} (\mu^2_f + \sigma^2_f - \log (\sigma_f) - 1), \nonumber \\
    \mathcal{L}_{rec} &= \mathcal{L}_{CE} + \mathcal{L}_{KL}, 
\end{align}
where $t_i^{(j)}$ is word $j$ in sample $i$, $n_f$ is the number of latent features assumed in the encoder, and $\mathcal{L}_{CE}$ and $\mathcal{L}_{KL}$ are the cross entropy and KL divergence loss functions in the news set $\mathcal{D}$. 
The classifier loss is given by:
%Variable $l_i$ denotes the length of post $i$, $V$ is the set of vocabulary defined by word2vec model, and $t_i^{(j)}$ is the $j$th word in post $i$. 
\begin{equation}
    \mathcal{L}_{BC} =  - \mathbb{E}_{i \sim D} \left[ y_i \log(\hat{y}_i) + (1-y_i) \log (1-\hat{y}_i) \right],
\end{equation}
where $y_i$ and $\hat{y}_i$ denote the label and the probability that news article $i$ is fake computed by the classifier. The optimized parameters $\theta^*_{VAE}$ minimize the total loss function:
\begin{equation}
\label{eq:overalobj}
    \theta^*_{VAE} = \text{argmin}_\theta (\mathcal{L}_{rec} + \mathcal{L}_{BC}),
\end{equation}

% The encoder is composed of two bidirectional LSTM (Bi-LSTM) and two fully connected layers. 
% Our motivation for using Bi-LSTM is that sequential layers often suffer from a problem of vanishing gradients which might lead to learning inefficient dependencies between the words by the model. LSTM layers, however, overcome this problem by including the forget gate and hidden states. 
% Furthermore, Bi-LSTM improves the accuracy over traditional LSTM, by utilizing two independent LSTM layers working side by side. The overview of the Bi-LSTM is depicted in Fig. \ref{figs:bilstm}.

% \begin{figure}[h]
% \centering 
% \includegraphics[width=0.5\textwidth]{Figs/Bilstm_one.PNG}
% \caption{Bi-LSTM}
% \label{figs:bilstm}
% \end{figure}

The output of the decoder is the generated news articles.
The decoder has a similar architecture to the encoder but in the opposite order of layers for the purpose of reconstructing the input from the latent space. 
% Together with the encoder and decoder, we train a 2-layered NN classifier. Our motivation for adding this part is obtaining more predictive feature space by including a supervised component to the model. 
In our framework, the desired features to be extracted from the VAE are the extracted latent features which are a $|\mathcal{D}| \times n_f$ matrix.

\subsection{Latent Dirichlet Allocation}
Latent Dirichlet allocation (LDA) is the unsupervised component in our model that jointly infers latent topics (distributions over words in a vocabulary) and the distribution of topics for each document. 
The generative model for LDA is as follows:
\begin{align}
     \beta_k & \sim \text{Dirichlet}_V(\bm{\eta}), \nonumber \\
     w_{ij} & \sim \text{Multinomial} (\beta_{z_{ij}}),\nonumber \\
     z_{ij} & \sim \text{Multinomial}(\varphi_i),\nonumber \\
     \varphi_i & \sim  \text{Dirichlet}_K(\bm{\alpha}).  \label{eq:ldagen}
\end{align}
We use stochastic variational inference to approximate the posterior distribution (Fig.~\ref{figs:graphicalModel} and Eq.~\ref{eq:joint})~\citep{hoffman2013stochastic}. 
In this context, stochastic variation inference uses stochastic optimization on the mean field variational distribution to iteratively optimize the evidence lower bound for the model specified in Equation~\ref{eq:ldagen}.
%Given the observed data $\bm{W}=(w_{ij})$, we maximize the likelihood function which is the best topic assignment of the words $\bm{Z}=(z_{ij})$ in the data:
%\begin{align}
%    \theta^*_{LDA} = \text{argmax}_\theta \text{ } p(\bm{Z}|\bm{W})
%\end{align}
Here, $w_{ij}$ is $j$\textsuperscript{th} word in the $i$\textsuperscript{th} post, and $z_{ij}$ determines the assignment of $w_{ij}$ to a topic.
The variational posterior for model parameter $\bm{\varphi} = (\varphi_i)$ is an $N \times K$ matrix where $N$ is the number of news articles, $K$ is the number of topics, and element $\varphi_{ik}$ represents the probability of generating a word from topic $\beta_k$ in news article $i$.
%In other words, $\varphi_{ik}$ is the proportion of words in that news $i$ has from topic $k$ ( and $k = {1, \dots, K}$).
The $K$-dimensional $\varphi_{i}$ vectors are concatenated to the corresponding $n_f$-dimensional latent features from the VAE (Fig.~\ref{figs:model}).

\subsection{Classifier}
The classifier receives the concatenated $\varphi_{i}$ and VAE embedding feature vector as input and outputs the news article label. 
We test six classifiers that include discriminative, generative, interpretable, and deep models: MLP, SVM, LR, Na\"ive Bayes (NB), RF, and KNN. 
Our motivation for including a diverse set of classifiers is to allow for model selection based on the characteristics of the data and desired downstream tasks (e.g., prioritizing interpretability of classification model). %to include different types of classifiers such as discriminative and generative classifiers is that, there is universally powerful classifier for all data.

\section{Results}
\label{sec:Results}

\subsection{Preprocessing}
Before the data were input into \texttt{LDAVAE}, the text was tokenized, non-English text and stop words were removed, hyperlinks, parentheses, and characters that are not expected to be in the words, such as `-', `@' and `\#' were removed. % TODO: ADD ALL PREPROCESSING STEPS
If the data is imbalanced, we over sample the minority class using a nearest neighbor extension of the Synthetic Minority Oversampling Technique (SMOTE) algorithm for categorical data (SMOTEN)~\citep{chawla2002smote}. % we could be precise with what we mean by "imbalance". 
These methods are commonly used for text classification in imbalanced data scenarios~\citep{zhao2021just}. 
% The preprocessing step includes SMOTEN only when the dataset is highly imbalanced.
Then as a preprocessing step for the VAE component, we transform the words to $w$-dimensional vectors by applying the distributed word2vec pre-trained model~\citep{mikolov2013distributed}. 
As a result of this transformation, semantically similar words are mapped closer to each other in the in the $w$ dimensional space than semantically dissimilar words.

\subsection{Datasets Studied}

We consider three fake news datasets that contain shorter OSM-shared news samples that we denote as \textit{news posts}.

% \section{Datasets}
% This section gives information about the datasets that we processed in this project.

\subsubsection{ISOT}
\begin{figure}[ht]
\centering 
\includegraphics[width=0.5\textwidth]{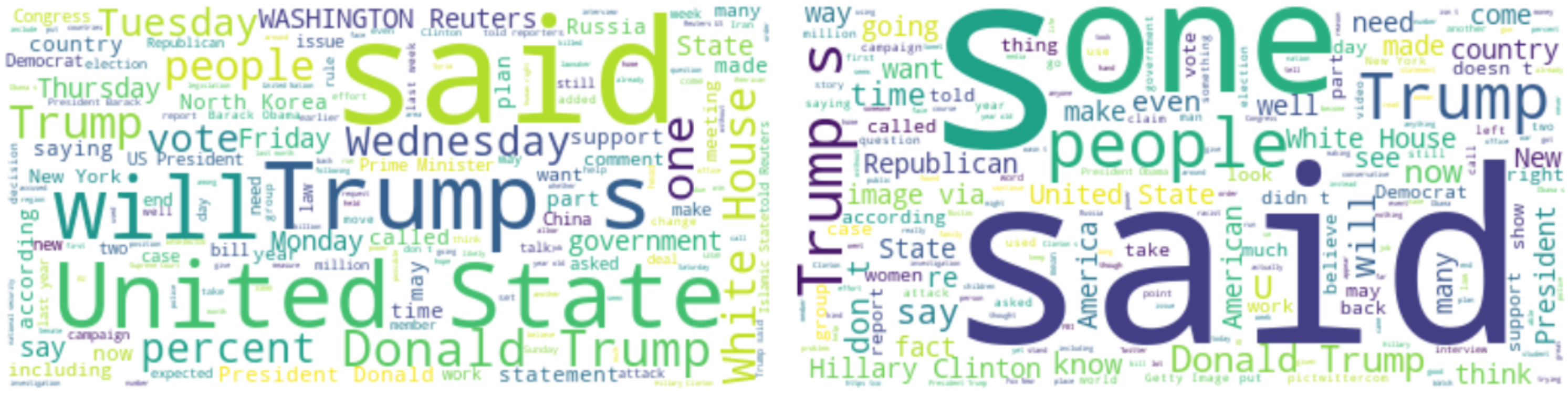}
\caption{\textbf{ISOT WordCloud} of ISOT real (left) and fake (right) data.}
\label{figs:data2wc}
\end{figure}

% The challenge regarding Twitter dataset is that the posts include both textual and visual content, and the text parts of the posts tend to be shorter than actual news. For this reason, we intend to use another dataset.
The ISOT fake news dataset
% ~\cite{ahmed2017detection, ahmed2018detecting} 
contains $44,898$ labeled news posts with a maximum length of $36$ words ($21,417$ real, $23,481$ fake)~\citep{ahmed2018detecting}. % was this before or after preprocessing? 
The authentic and fake news are collected from the news agency Reuters and unreliable websites (flagged by Politifact) respectively. The majority of the data are political news from $2016$ to $2017$.
Although the data are cleaned, the punctuation and grammatical mistakes in the fake news posts are retained in the original data~\citep{ahmed2018detecting}. 
After preprocessing, the number of samples was reduced to $44,143$ ($22,727$ fake and $21,416$ real). 
%\footnote{\href{https://www.uvic.ca/engineering/ece/isot/datasets/}{https://www.uvic.ca/engineering/ece/isot/datasets/}}.
The similarity between real and fake news post word usage in the ISOT dataset suggests the real and fake news posts are not well separated by word frequencies alone (Fig.~\ref{figs:data2wc}). 

\subsubsection{COVID}

\begin{figure}[ht]
\centering 
\includegraphics[trim=140 180 110 180,clip, width=0.49\textwidth]{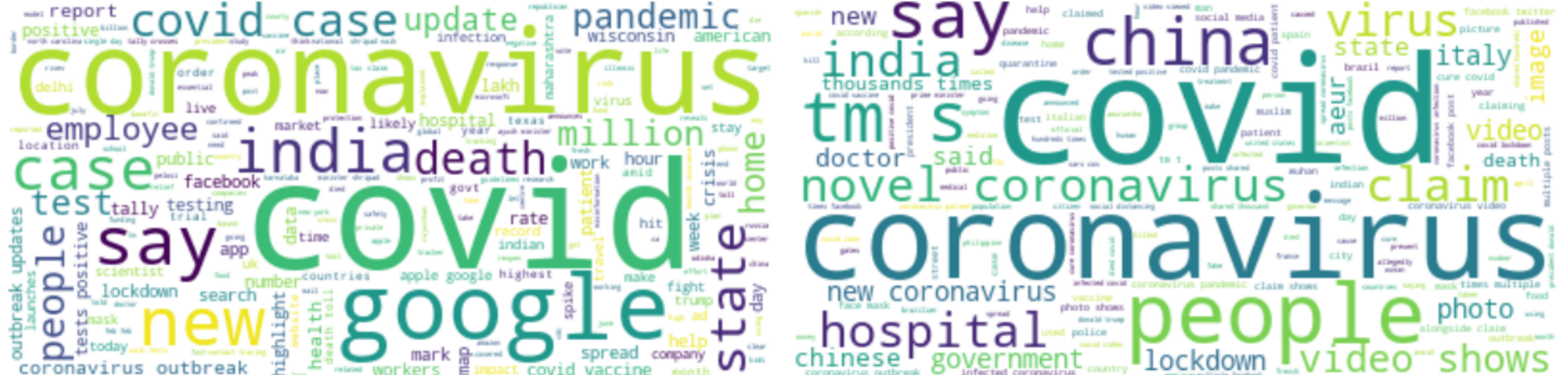}
\caption{\textbf{COVID WordCloud} of real (left) and fake (right) data.}
\label{figs:Covidwc}
\end{figure}

Published in Nov 2020, the COVID dataset is a recently collected and labeled dataset that contains $10,201$ news headlines shared on OSM related to COVID-19~\citep{covid_data}. 
%It was published on November 2020 and combines other datasets for fake news detection. 
After preprocessing, the COVID data consisted of $9,999$ 
%($9536$ fake, $463$ real)
news headlines with binary real or fake labels, and a maximum post length of $115$ words. 
% The original data has $10,201$ posts, with $9727$ fake and $474$ real labels.
Classifying fake news headlines is challenging in this data due to class imbalance. 
Among the $9,999$ headlines, only $463$ are labeled as real news, so we synthesized $6,695$ additional real news samples for a total sample size of $16,694$.
Unlike the ISOT data, word usage appears to be qualitatively different between real and fake news posts; e.g., terms associated with China are more prevalent in fake news posts (Fig.~\ref{figs:Covidwc}). 
%This dataset is publicly accessible~\cite{covid_data}.
% \footnote{\href{https://zenodo.org/record/4282522\#.YKlJXahKjtU}{https://zenodo.org/record/4282522\#.YKlJXahKjtU}}.

\subsubsection{Twitter}
\begin{figure}[ht]
\centering 
\includegraphics[trim=140 180 110 180,clip, width=0.49\textwidth]{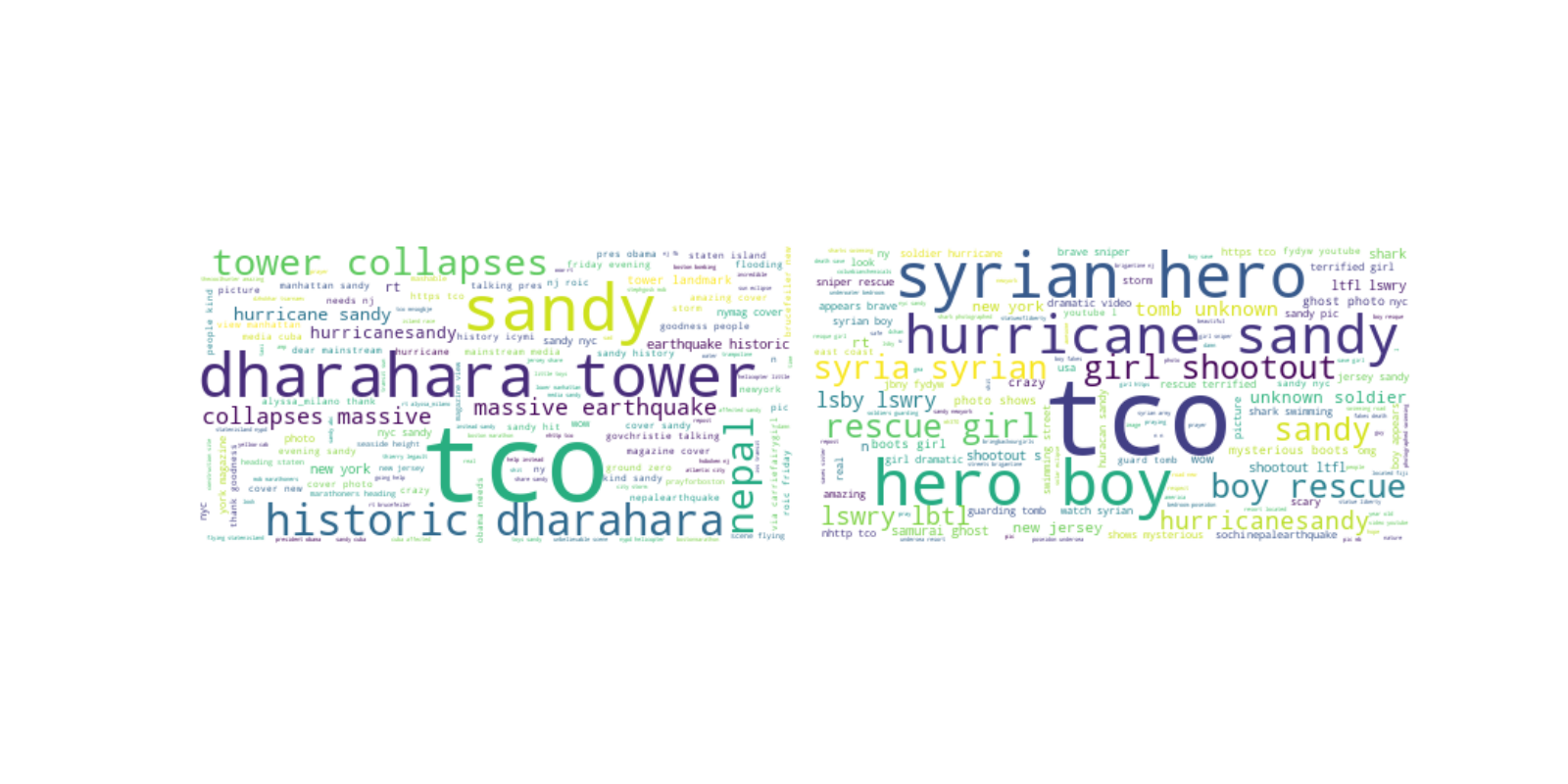}
\caption{\textbf{Twitter WordCloud} of real (left) and fake (right) data.}
\label{figs:data1wc}
\end{figure}

The Twitter dataset~\citep{boididou2018detection} is another benchmark that was originally collected for the MediaEval Workshop in 2016~\citep{twitter_data}.
% \footnote{\href{https://github.com/MKLab-ITI/image-verification-corpus}{https://github.com/MKLab-ITI/image-verification-corpus}}. 
It includes $15,629$ labeled tweets of maximum length $31$ words and covering $17$ events. The number of fake and real posts are $9,404$ and $6,225$ respectively. 
After preprocessing, the number of samples was reduced to $12,242$, among which $7,205$ are fake and $5,037$ are real. 
The data contains information about the news post such as the source and textual content. 
Our motivation to use this dataset is three-fold: it is often used to evaluate fake news classifiers~\citep{can2019new, khattar2019mvae, verdoliva2020media}; it contains tweets from different events; and we expect the LDA model will find distinguishable topics after training. 
However, this dataset is challenging due to the small length of news posts and the presence of noise and many non meaningful words in the text, which is reflected in the word frequencies (Fig.~\ref{figs:data1wc}).
%Additionally, the news contain textual and visual content and the textual part is short (the longest tweet contains 29 words). 
% subplots correspond to training and test sets respectively.
%Fig. \ref{figs:data1wc} shows the WordClouds of real and fake tweets after prepeocessing.

% \vspace{-0.3cm}
\subsection{Evaluation criteria}

% \hl{list comparison schemes in bullets}
% \begin{itemize}
%     \item 
%     \item
% \end{itemize}

% \hl{list comparison metrics in bullets}

We evaluate the performance of \texttt{LDAVAE} both qualitatively and quantitatively and also internally with respect to the different input feature sets and classifier models and externally to $4$ state-of-the-art competing methods.

\subsubsection{Interpreting model parameters}
We qualitatively evaluate the separability of features sets obtained by the VAE, LDA, and their concatenation on training and test sets using PCA and t-SNE. 
The concatenated feature matrices of size $|\mathcal{D}_{tr}| \times (n_f + K)$ and $|\mathcal{D}_{te}| \times (n_f + K)$ for $\mathcal{D}_{tr}$ and $\mathcal{D}_{te}$ respectively, are embedded down to $2$ dimensions. 
% The motivation for applying dimensionality reduction is to illustrate the concatenation of two feature sets can improve the separation of different classes. %?
% This is particularly useful in unsupervised settings or if the labels of news articles are not available.
We also consider the distribution of topics across fake and real articles.
We visualize the average posterior mean of the variational parameters for $\varphi_i$ across real and fake news articles with radar plots. 

\subsubsection{Metrics on classification outcome}
We compute accuracy, F1 score, false positive rate, and false negative rates internally for \texttt{LDAVAE} using the feature sets obtained by VAE, LDA, and their concatenation. 
Accuracy is the ratio of the total correct labels to the size of the dataset:
\begin{equation*}
\text{Accuracy} = \frac{\text{TP + TN}}{\text{TP + TN + FP + FN}},
\label{eg:Accuracy}
\end{equation*}
where TP, TN, FP, and FN are true positives, true negative, false positives, and false negatives, respectively.
Precision is the fraction of actual fake news among all the fake detected news $\frac{\text{TP}}{\text{TP + FP}}$ and recall is the ratio of news truly detected as fake to all the fake news in the data $\frac{\text{TP}}{\text{TP + FN}}$.
We present the precision and recall as their harmonic mean: 
\begin{equation*}
    \text{F1 score} = 2 \times \frac{\text{Precision} \times \text{Recall}}{\text{Precision + Recall}},
\end{equation*}
We also consider the false positive ($\text{FPR} = \frac{\text{FP}}{\text{TN + FP}}$) and false negative rates ($\text{FNR} = \frac{\text{FN}}{\text{FN + TP}}$).

\subsection{Experimental Settings}

%We apply our method on three datasets. 
% Then we filter out non-English posts using langdetect Python package.
% In the next step, 
Each news post is considered as a vector of $L$ words $t_i^{(1)}, \dots, t_i^{(l_i)}$ where $L = max\{l_i:i = 1,\dots, N\}$. 
Using a word2vec model pretrained on the Google News dataset~\citep{rehurek2011gensim,mikolov2013distributed}, we produced $w$-dimensional embeddings for each word in the three datasets resulting in a $|V| \times w$ embedding matrix, where $|V|$ is the size of vocabulary set in the model.
%provided in gensim package (version 3.x or 4.x) on the dataset $\mathcal{D}$, 
%and obtain a $w$-dimensional vector for each vocabulary $v \in V$ detected by word2vec. 
%The result is a 
%Depending on $w$, $|V|$ might change. 
% Then we split ISOT dataset $\mathcal{D}$ into training and test sets ($\mathcal{D}_{tr}$ and $\mathcal{D}_{te}$), with 20\% of the data as test and stratification of the news label. 
% In this project, for different datasets, we have set $w \in \{8, 16, 32, 64, 160\}$. Twitter dataset has separated training and test sets.
The input to the VAE is the concatenation of all word embeddings in a news post; if $l_i \leq L$, we apply zero-padding to set the input vector dimension to $L\times w$.  
For each dataset, the VAE is trained to minimize the reconstruction and classifier loss (Eq.~\ref{eq:overalobj}) on $90\%$ of the data. 
Architectural details, code documentation, and details for reproducing the subsequent analyses are available in the code repository\footnote{\url{https://github.com/Marjan-Hosseini/LDAVAE}}. 
LDA is trained on the same $90\%$ training data as the VAE, however we make the assumption that a news post can be represented as a set of words in the vocabulary $V$ (\emph{i.e.} bags of words). 
%We vectorize using tf-idf and let the maximum number of the allowed vocabulary be greater than $|V|$ ($=$ 20000).
%We use the same vocabulary set that was input to the word2vec model.
%Moreover, LDA is a parametric model and the number of topics ($K$) should be set in advance. 
% We change $K \in \{8, 10, 16, 32, 64\}$. In addition, we chose the number of iterations in sampling $=1000$.
We tune the number of topics $K \in \{8, 10, 16, 32, 64\}$ using the coherence score of the trained model on a validation set. 
Although perplexity and predictive likelihood are also used for model evaluation, maximizing coherence has been previously associated with better interpretability \citep{roder2015exploring}. % proven? as in there was a theoretical proof? I don't think so...
% \subsection{Classifier}
We concatenate the features extracted from the LDA and VAE models and apply six widely used classifiers, MLP, RF, SVM, LR, NB, and KNN.
% Then, we compute PCA and t-SNE for visualization.
% Then, we compute PCA and t-SNE dimensionality reduction for visualization.
%from scikit-learn Python package with the default setting. For MLP we set the number of iterations equal to 300. By default, Random Forest classifier has the number of estimators equal to 100. In SVM we use the linear kernel, and in KNN, we select k = 3. In logistic regression classifier, we use liblinear solver. Na\"ive Bayes has the default setting. Before applying all the classifiers, we normalize the features with z-Score normalization using StandardScaler in scikit-learn package. 
% \subsection{Evaluation}
% We computed PCA and t-SNE dimensionality reduction on features obtained by VAE, LDA, and their concatenation. In both methods, we selected the number of components $=2$ for better visualization. In t-SNE, we set perplexity $=40$. For PCA, we normalize data with z-score normalization before transformation, to obtain eigenvalues from the features in the same scale. 
% Then the classifiers are trained using the training set and the entire feature sets, and all the metrics provided in Sect.~\ref{sec:syseval} are computed for $\mathcal{D}_{tr}$ and $\mathcal{D}_{te}$. In the next experiment, we first apply feature selection and after that, we repeat the classification. 

% \section{Results}
%\subsection{Experimental Results}
%\label{sec:results}
%This section includes results ...

\subsection{Fake News Classification Results}

\subsubsection{Ablation Study}
We first evaluated the usefulness of integrating neural embeddings and topic features in \texttt{LDAVAE} by comparing the news post classification accuracy using only LDA features, only VAE features, and combined LDA and VAE features.
%Besides the model deep classifier,  
Simultaneously, we compare the accuracy, F1 score, FPR, and FNR for the six classifiers across the three datasets.
For most of the dataset and evaluation criteria configurations, classifier performance was higher when the two feature sets are concatenated (Tab.~\ref{tab:metrics_all}). 
% Here, as expected, LDA features do not show predictive results compared to VAE, since the setting for LDA is supervised, while VAE is coupled with a classifier. However,
%Even if the evaluation metrics for LDA features alone are not strinctly better than VAE, their concatenation seems to improve the accuracy metrics in most of the classifiers. 
While the \texttt{LDAVAE} is most frequently the highest scoring model configuration, topic features alone (LDA) achieve a higher accuracy and F1 score in the Twitter dataset, potentially because the frequency of fake news posts vary based on topics~\citep{torabi2019big}.
In fact, the VAE only classifiers perform poorly on the Twitter dataset in general possibly due to the short and unstructured nature of Twitter news posts.
This underscores the usefulness of incorporating topic features into fake news classification.
Since random forest classifiers yielded the best aggregate results, we only consider random forest classifiers for the subsequent results.
% Moreover, 
% the results in this section and  Tab.~\ref{tab:comparison} show our method has reduce the overfitting. 

\begin{table}[h] 
\caption{Comparison of the accuracy metrics after running the model on ISOT, COVID, and Twitter datasets and classifying VAE, LDA feature set (baseline), and their concatenation. The best scores are shown in bold.} 
\label{tab:metrics_all} 
\resizebox{0.48\textwidth}{!}{
\begin{tabular}{ll|ccc|ccc|ccc|} 
\cline{3-11} 
&  & \multicolumn{3}{c|}{ISOT} & \multicolumn{3}{c|}{COVID} & \multicolumn{3}{c|}{Twitter} \\ \cline{3-11} 
 & {\color[HTML]{000000} } & \multicolumn{1}{c|}{{\color[HTML]{000000} VAE}} & \multicolumn{1}{c|}{{\color[HTML]{000000} LDA}} & \multicolumn{1}{c|}{{\color[HTML]{000000} \texttt{LDAVAE}}} & \multicolumn{1}{c|}{VAE} & \multicolumn{1}{c|}{LDA} & \multicolumn{1}{c|}{\texttt{LDAVAE}} & \multicolumn{1}{c|}{VAE} & \multicolumn{1}{c|}{LDA} & \multicolumn{1}{c|}{\texttt{LDAVAE}} \\ \hline 
\multicolumn{1}{|c|}{} & Acc. & \multicolumn{1}{c}{0.88} &\multicolumn{1}{c}{0.85} &\multicolumn{1}{c|}{0.89} & 0.90 & 0.62 & 0.90 & 0.74 & 0.81 & 0.82 \\
\multicolumn{1}{|c|}{} & F-Sc. & \multicolumn{1}{c}{0.87} &\multicolumn{1}{c}{0.83} &\multicolumn{1}{c|}{0.88} & 0.90 & 0.41 & 0.91 & 0.59 & 0.71 & 0.70 \\
\multicolumn{1}{|l|}{\multirow{-2}{*}{SVM}} & FPR & 0.13 & 0.13 & 0.11 & 0.18 & \textbf{0.01} & 0.16 & 0.16 & 0.12 & 0.09 \\
\multicolumn{1}{|l|}{} & FNR &0.11 & 0.17 & 0.10 & 0.03 & 0.74 & 0.03 & 0.41 & 0.29 & 0.30 \\ \hline 
\multicolumn{1}{|c|}{} & Acc. & \multicolumn{1}{c}{0.88} &\multicolumn{1}{c}{0.87} &\multicolumn{1}{c|}{0.89} & 0.91 & 0.61 & 0.91 & 0.73 & 0.79 & 0.79 \\
\multicolumn{1}{|c|}{} & F-Sc. & \multicolumn{1}{c}{0.87} &\multicolumn{1}{c}{0.85} &\multicolumn{1}{c|}{0.88} & 0.91 & 0.40 & 0.91 & 0.59 & 0.70 & 0.72 \\
\multicolumn{1}{|c|}{\multirow{-2}{*}{LR}} & FPR & 0.14 & 0.13 & 0.11 & 0.15 & 0.03 & 0.15 & 0.18 & 0.15 & 0.17 \\
\multicolumn{1}{|c|}{} & FNR &0.10 & 0.14 & 0.10 & 0.03 & 0.74 & 0.03 & 0.41 & 0.30 & 0.28 \\ \hline 
\multicolumn{1}{|c|}{} & Acc. & \multicolumn{1}{c}{0.88} &\multicolumn{1}{c}{0.88} &\multicolumn{1}{c|}{\textbf{0.90}} & 0.90 & 0.58 & \textbf{0.92} & 0.73 & \textbf{0.84} & 0.82 \\
\multicolumn{1}{|l|}{} & F-Sc. & \multicolumn{1}{c}{0.87} &\multicolumn{1}{c}{0.87} &\multicolumn{1}{c|}{\textbf{0.89}} & 0.91 & 0.44 & \textbf{0.92} & 0.65 & \textbf{0.75} & 0.74 \\
\multicolumn{1}{|c|}{\multirow{-2}{*}{RF}} & FPR & 0.13 & 0.10 & \textbf{0.10} & 0.15 & 0.17 & 0.10 & 0.21 & \textbf{0.09} & 0.13 \\
\multicolumn{1}{|c|}{} & FNR &0.11 & 0.14 & \textbf{0.08} & 0.05 & 0.67 & 0.06 & 0.35 & \textbf{0.25} & 0.26 \\ \hline 
\multicolumn{1}{|c|}{} & Acc. & \multicolumn{1}{c}{0.88} &\multicolumn{1}{c}{0.83} &\multicolumn{1}{c|}{0.88} & 0.90 & 0.51 & 0.51 & 0.74 & 0.77 & 0.76 \\
\multicolumn{1}{|c|}{} & F-Sc. & \multicolumn{1}{c}{0.87} &\multicolumn{1}{c}{0.81} &\multicolumn{1}{c|}{0.87} & 0.91 & 0.67 & 0.67 & 0.78 & 0.74 & 0.81 \\
\multicolumn{1}{|c|}{\multirow{-2}{*}{NB}} & FPR & 0.15 & 0.18 & 0.14 & 0.16 & 0.97 & 0.97 & 0.29 & 0.21 & 0.28 \\
\multicolumn{1}{|c|}{} & FNR &0.09 & 0.17 & 0.09 & 0.03 & \textbf{0.00} & \textbf{0.00} & 0.22 & 0.26 & 0.19 \\ \hline 
\multicolumn{1}{|c|}{} & Acc. & \multicolumn{1}{c}{0.88} &\multicolumn{1}{c}{0.86} &\multicolumn{1}{c|}{0.89} & 0.89 & 0.58 & 0.91 & 0.74 & 0.78 & 0.82 \\
\multicolumn{1}{|c|}{} & F-Sc. & \multicolumn{1}{c}{0.87} &\multicolumn{1}{c}{0.85} &\multicolumn{1}{c|}{0.87} & 0.90 & 0.53 & 0.91 & 0.66 & 0.73 & 0.72 \\
\multicolumn{1}{|c|}{\multirow{-2}{*}{MLP}} & FPR & 0.13 & 0.13 &\textbf{0.10} & 0.18 & 0.31 & 0.12 & 0.21 & 0.18 & 0.11 \\
\multicolumn{1}{|c|}{} & FNR &0.11 & 0.14 & 0.13 & 0.03 & 0.53 & 0.06 & 0.34 & 0.27 & 0.28 \\ \hline 
\multicolumn{1}{|c|}{} & Acc. & \multicolumn{1}{c}{0.85} &\multicolumn{1}{c}{0.80} &\multicolumn{1}{c|}{0.88} & 0.86 & 0.59 & 0.85 & 0.73 & 0.77 & 0.82 \\
\multicolumn{1}{|c|}{} & F-Sc. & \multicolumn{1}{c}{0.83} &\multicolumn{1}{c}{0.77} &\multicolumn{1}{c|}{0.87} & 0.87 & 0.44 & 0.86 & 0.67 & 0.71 & 0.74 \\
\multicolumn{1}{|c|}{\multirow{-2}{*}{KNN}} & FPR & 0.14 & 0.16 & \textbf{0.10} & 0.21 & 0.14 & 0.21 & 0.23 & 0.19 & 0.13 \\
\multicolumn{1}{|l|}{} & FNR &0.16 & 0.24 & 0.14 & 0.07 & 0.68 & 0.08 & 0.33 & 0.29 & 0.26 \\ \hline 
\end{tabular} }
\end{table}

\subsubsection{Comparison with SOTA}
Next, we compare the performance of \texttt{LDAVAE} to four SOTA methods in the same category of fake news classification (content-based): MVAE~\citep{khattar2019mvae}, FakeDetector~\citep{zhang2020fakedetector}, Hybrid CNN~\citep{wang2017liar} and RNN~\citep{mikolov2010recurrent}.
%These methods were previously being compared by the FakeDetector method which is  similar to our framework. 
To keep the analysis consistent across methods, we modified the methods to only consider textual data, e.g., we removed the layers of MVAE associated with image processing.
\begin{table}[h]
\centering
\caption{Comparison of the metrics among different methods (an RF classifier is used with \texttt{LDAVAE}).}
\label{tab:comparison}
\resizebox{0.48\textwidth}{!}{%
\begin{tabular}{l|l|l|l|l|}
\cline{2-5}
                                   & \multicolumn{1}{c|}{\multirow{2}{*}{Metric}} & \multicolumn{1}{c|}{ISOT}         & \multicolumn{1}{c|}{COVID}        & \multicolumn{1}{c|}{Twitter}      \\
                                   & \multicolumn{1}{c|}{}                        & \multicolumn{1}{c|}{Train (Test)} & \multicolumn{1}{c|}{Train (Test)} & \multicolumn{1}{c|}{Train (Test)} \\ \hline
\multicolumn{1}{|l|}{\texttt{LDAVAE}}         & Acc.                                         &     1.00 (\textbf{0.90})	               &       1.00 (0.92)                &   		1.00 (0.82)              \\
\multicolumn{1}{|l|}{}             & F-Sc.                                        &     1.00 (0.89)                  &       1.00 (0.92)                &         0.75 (0.74)              \\
\multicolumn{1}{|l|}{}             & FPR                                          &     0.00 (0.10)                  &       0.00 (0.10)                &         0.08 (0.13)              \\
\multicolumn{1}{|l|}{}             & FNR                                          &     0.00 (0.08)                  &       0.00 (0.06)                &         0.24 (0.26)              \\ \hline
\multicolumn{1}{|l|}{MVAE~\citep{khattar2019mvae}}         & Acc.                  &     0.87 (0.88)                  &       0.78 (0.79)                &         0.76 (0.73)              \\
\multicolumn{1}{|l|}{}             & F-Sc.                                        &     0.89 (0.89)                  &       0.07 (0.08)                &         0.63 (0.59)              \\
\multicolumn{1}{|l|}{}             & FPR                                          &     0.13 (0.12)                  &       0.18 (0.17)                &         0.14 (0.16)              \\
\multicolumn{1}{|l|}{}             & FNR                                          &     0.11 (0.11)                  &       0.83 (0.80)                &         0.37 (0.40)              \\ \hline
\multicolumn{1}{|l|}{FakeDetector~\citep{zhang2020fakedetector}} & Acc.            &     0.94 (0.82)                  &       0.90 (0.92)                &         0.94 (0.85)              \\
\multicolumn{1}{|l|}{}             & F-Sc.                                        &     0.94 (0.81)                  &       0.91 (0.92)                &         0.93 (0.82)               \\
\multicolumn{1}{|l|}{}             & FPR                                          &     0.06 (0.07)                  &       0.09 (0.13)                &         0.04 (0.12)              \\
\multicolumn{1}{|l|}{}             & FNR                                          &     0.04 (0.07)                  &       0.09 (0.03)                &         0.07 (0.17)              \\ \hline
\multicolumn{1}{|l|}{Hybrid CNN~\citep{wang2017liar}}    & Acc.                     &     0.93 (0.88)                  &       0.95 (\textbf{0.95})                &         0.90 (0.86)              \\
\multicolumn{1}{|l|}{}             & F-Sc.                                        &     0.93 (0.87)                  &       0.95 (0.95)                &         0.90 (0.79)              \\
\multicolumn{1}{|l|}{}             & FPR                                          &     0.07 (0.10)                  &       0.04 (0.04)                 &         0.04 (0.10)              \\
\multicolumn{1}{|l|}{}             & FNR                                          &     0.06 (0.12)                  &       0.04 (0.04)                &         0.10 (0.21)              \\ \hline
\multicolumn{1}{|l|}{RNN~\citep{mikolov2010recurrent}}  & Acc.                     &     0.75 (0.74)                  &       0.99 (0.78)                 &         0.98 (\textbf{0.89})              \\
\multicolumn{1}{|l|}{}             & F-Sc.                                        &     0.49 (0.47)                  &       0.99 (0.75)                 &         0.97 (0.84)              \\
\multicolumn{1}{|l|}{}             & FPR                                          &     0.03 (0.03)                  &       0.02 (0.10)                &         0.01 (0.08)              \\
\multicolumn{1}{|l|}{}             & FNR                                          &     0.51 (0.52)                  &       0.01 (0.35)                 &         0.03 (0.15)              \\ \hline
\end{tabular}%
}
\end{table}

\texttt{LDAVAE} achieves the best accuracy and F1 score in the ISOT dataset, which is the largest dataset in terms of news post quantity and is also well-curated compared to the other datasets since it includes articles from Reuters and human labeled fake news posts; it also achieves comparable results on the COVID and Twitter data (Tab.~\ref{tab:comparison}).
Both the Twitter and COVID datasets are noisy and model training was performed without syntactic or semantic text correction.
%Consequently, the text requires extensive and dataset-specific pre-processing. 
%However, to be fair we did not change our pre-processing for these dataset. 
Among the three datasets, the performance of VAE-based methods is worst on the Twitter data likely due to having few samples and news posts having short length. % which is a known limitation for topic models~\cite{chen2016short}  <-- we can't say this because Twitter has the shorted length but LDA does the best for twitter!
Additionally, the VAE objective function, which requires balancing the reconstruction error, classification error, and sample generation quality, is difficult to balance during training for specific downstream tasks such as classification~\citep{bohm2020probabilistic}. 
%FakeDetector and Hybrid CNN do not suffer this problem, however, MVAE does, which is reflected in the results.
%Among VAE based methods, our architecture clearly outperforms on all but F1 score for ISOT where the methods are tied.
% MVAE does not outperform our method in any of the datasets.
RNNs use a simpler autoregressive architecture and the results deteriorate when the sequence length is increased. 
%The stated problem can be tackled using Normalizing Flows (NF)~\cite{papamakarios2019normalizing} that optimizes the objective functions one by one. % what???

\subsection{Model Interpretability}
Here, we explore the interpretability of \texttt{LDAVAE} through analysis and visualization of the parameters in the LDA model.
% \begin{figure*}[h]

%   \begin{subfigure}{\linewidth}
%   \includegraphics[trim=170 0 200 10,clip,width=1\linewidth]{Figs/ISOT_lda_topic_tf_5.pdf}
%   \end{subfigure}\par\medskip
%   \begin{subfigure}{\linewidth}
%   \includegraphics[trim=170 0 200 10,clip,width=1\linewidth]{Figs/lda_topic_tf_5_Covid.pdf}
%   \end{subfigure}\par\medskip
%   \begin{subfigure}{\linewidth}
%   \includegraphics[trim=170 0 200 10,clip,width=1\linewidth]{Figs/Twitter_lda_topic_tf_5.pdf}
%   \end{subfigure}\par\medskip
%   \caption{The most frequent words in the extracted topics in ISOT, Covid and Twitter datasets (top to bottom).}
%   \label{fig:allInOneFreq}
% \end{figure*}

% \begin{figure*}[h]
% \centering 
% \includegraphics[width=1\textwidth]{Figs/allInOne.pdf}
% \caption{The most frequent words in the extracted topics in ISOT, Covid and Twitter datasets (top to bottom).}
% \label{fig:allInOneFreq}
% \end{figure*}

\subsubsection{News Post Topic Distributions}
% Unlike many of the existing methods in the literature that include human supervised, dataset-dependent subjects extraction in text processing/classification, we find the topics automatically using LDA model. 
% Fig.~\ref{fig:allInOneFreq} presents
% % Fig.~\ref{figs:isotfreq},~\ref{figs:Covidfreq}, and~\ref{figs:ldafreqtwitter} are 
% the most frequent words in the obtained topics in all datasets. 

\begin{figure*}[h, align=\centering, width=1\textwidth]
\centering 
\includegraphics[width=0.981\textwidth]{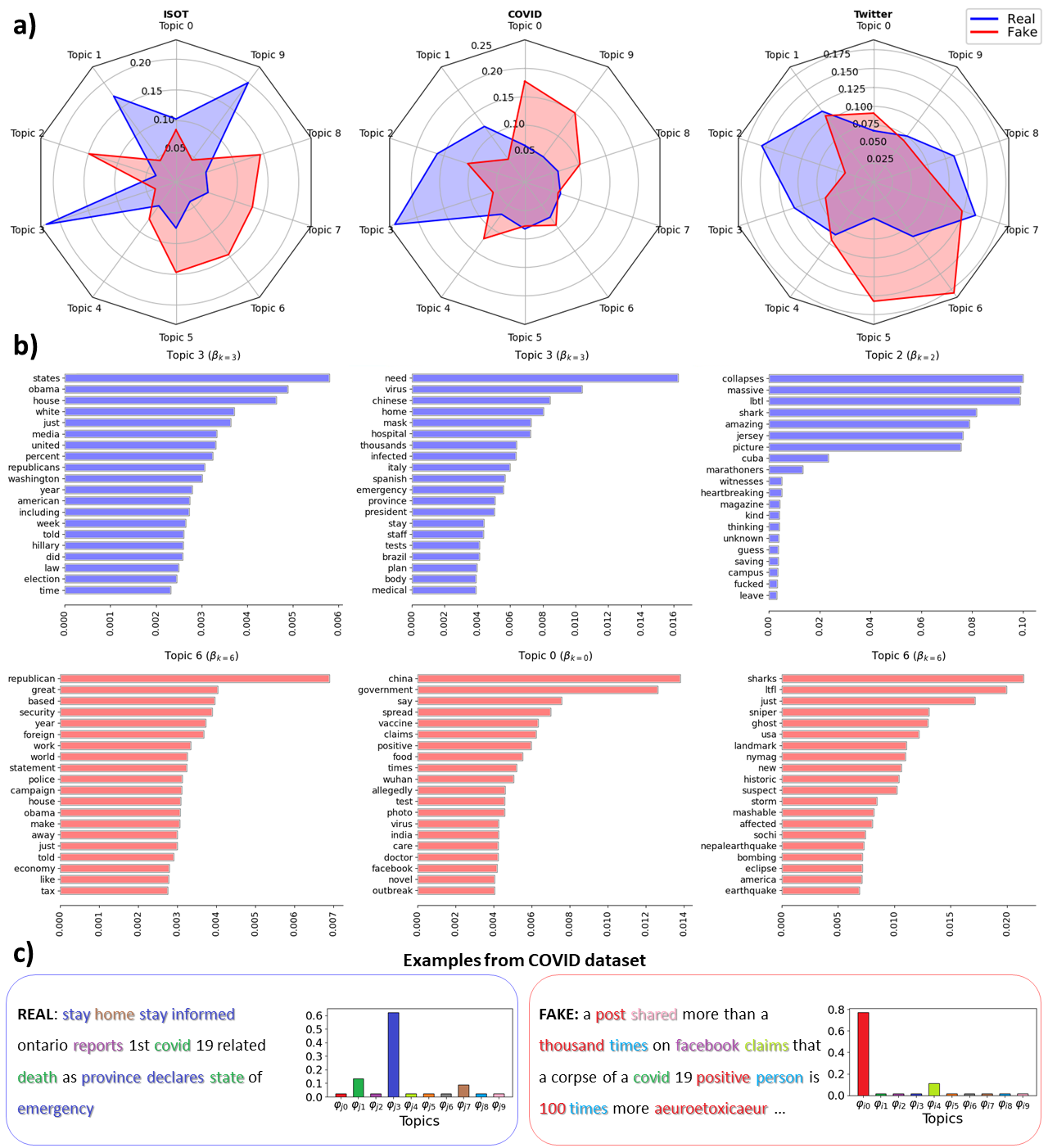}
\caption{\textbf{Model Interpretability.} a) Averaged posterior means of $\varphi_i$ (per news post topic distribution) separated by their labels. b) Normalized frequency of the top 20 most frequent words in the two topics where the difference in the average posterior means of $\varphi_i$ variables between real (blue) and fake news (red) posts is the largest (i.e. two most discriminative topics in each dataset). The word distributions for each dataset are shown below the associated radar plots. c) Interpreting two example sentences selected from the COVID dataset. 
We show a real sentence example ($j$) and a fake sentence example ($i$), along with the plot of posterior means of $\varphi$ variables ($\varphi_j$ and $\varphi_i$); each bar represents a topic in $\varphi$, and we colored the words in the sentences based on their most probable topic assignment ($z_{j.}$ and $z_{i.}$ where the dot represents the vector of words in the sentence). 
}
\label{figs:radar_with_cov_ex}
\end{figure*}

The variable posteriors of $\beta_k$ and $\varphi_i$ are interpretable as the distribution of words in topics and the distribution of topics in the news posts respectively. 
Further, posterior inference yields a distribution rather than point estimates, consequently the parameter uncertainty can be presented to users to give flexibility in interpreting variable importance and effect regarding the class label. 
For example, we 
interpret the effect of topics on class outcome by visualizing the topic distribution across news posts and separated by their labels. 
Specifically, we set the number of topics $K=10$ (for visualization purposes) and plot the averaged posterior means of $\varphi_i$ for real and fake news posts (Fig.~\ref{figs:radar_with_cov_ex}).
% This corresponds to the distribution of topics in the fake and real news articles. 
% Subject credibility has been previously computed in FakeDetector~\cite{zhang2020fakedetector} for manual topics.?
The frequencies of topics in real and fake news posts are largely distinct with some topic overlap. % different and the weights of words associated with those topics can be easily inspected and interpreted by posterior of $\beta_k$ variables.
% If topics that are more discriminative can be interpreted as large differences in average posterior means, then clearly the topics are the least informative in the Twitter data, where only $2$ of the $10$ topics appear to have discriminative power.
The topics that are more discriminative can be interpreted as larger differences in average posterior means. 
For example, all topics besides $0$ and $4$ are informative in the ISOT dataset (Fig.~\ref{figs:radar_with_cov_ex} (a)).
%In Fig.~\ref{figs:radar_with_cov_ex} (b) We plotted top words for $\beta_k$ in two top most informative topics for all three datasets. Then in Fig.~\ref{figs:radar_with_cov_ex} (b)

% we need to be *exact* with our language. What does "topic distribution" mean? The model parameters are *distributions* which means in order to plot them on a radar plot like this, you need to perform some processing. I talked with marjan and I think you took the average of the posterior means.

%($z_{j.}$ and $z_{i.}$ where dot represent the vector of words in the sentence).

% , with interpretation according to \bm{\beta} and \bm{\varphi} % Averaged posterior means of $\varphi_i$ shows differences in real and fake news post topic usage across the ISOT (left), Covid (middle), and Twitter (right) datasets

% \begin{figure*}[]
% \centering 
% \includegraphics[width=1\textwidth]{Figs/radar_all.jpg}
% \caption{\textbf{Interpreting $\bm{\varphi}$.} Averaged posterior means of $\varphi_i$ shows differences in real and fake news post topic usage across the ISOT (left), Covid (middle), and Twitter (right) datasets.}
% \label{figs:subjects_radar}
% \end{figure*}

\subsubsection{Topic Distributions}
For each dataset, we compute the most discriminatory topics using \texttt{LDAVAE} parameters $\bm{\varphi}$ and $\bm{\beta}$.
We consider the two topics that have the largest difference in the average posterior means of $\varphi_i$ variables across real and fake news posts.
In other words, we sort the vector $(|\sum_{i \in \mathcal{D}_{tr}^F} \hat{\varphi_{ik}}-\sum_{i \in \mathcal{D}_{tr}^R} \hat{\varphi_{ik}}|)_{k=1}^K$ in descending order where $\mathcal{D}_{tr}^F$ and $\mathcal{D}_{tr}^R$ are the training sample sets for fake and real news posts respectively.
We then plot the normalized frequency of the top $20$ most frequent words~(Fig.~\ref{figs:radar_with_cov_ex} (b, c)). % \break %this break puts some awkward space at the bottom of the page. If you think we need it though, you can uncomment
Word distributions in discriminative topics are subtly different.
For example, while words associated with U.S. conservative politics and China were found in both topics associated with real and fake news, their frequencies were substantially different. % if you see anything else interesting, please feel free to add

% \hl{RESULT IDEA:  credibility of different topics, but we actually infer the topics unlike https://arxiv.org/pdf/1805.08751.pdf (figure f)}

% \hl{RESULT IDEA: plot the $\theta$ variable (distribution over topics for each document) for the fake vs the real articles. This is similar to a plot Marjan made for isoforms. We sort by the percent in a specific topic. This was done for population distributions in humans in the fastSTRUCTURE paper.}

% \hl{RESULT IDEA: Try showing 2 t-SNE plots for the same set of points, one colored by fake/real and the other colored by the topic of highest frequency in the document. This might show that the topics help us differentiate real and fake.}

\subsubsection{Dimensionality Reduction}

\begin{figure}[h]
\centering 
\includegraphics[trim=10 10 0 10,clip,width=0.50\textwidth]{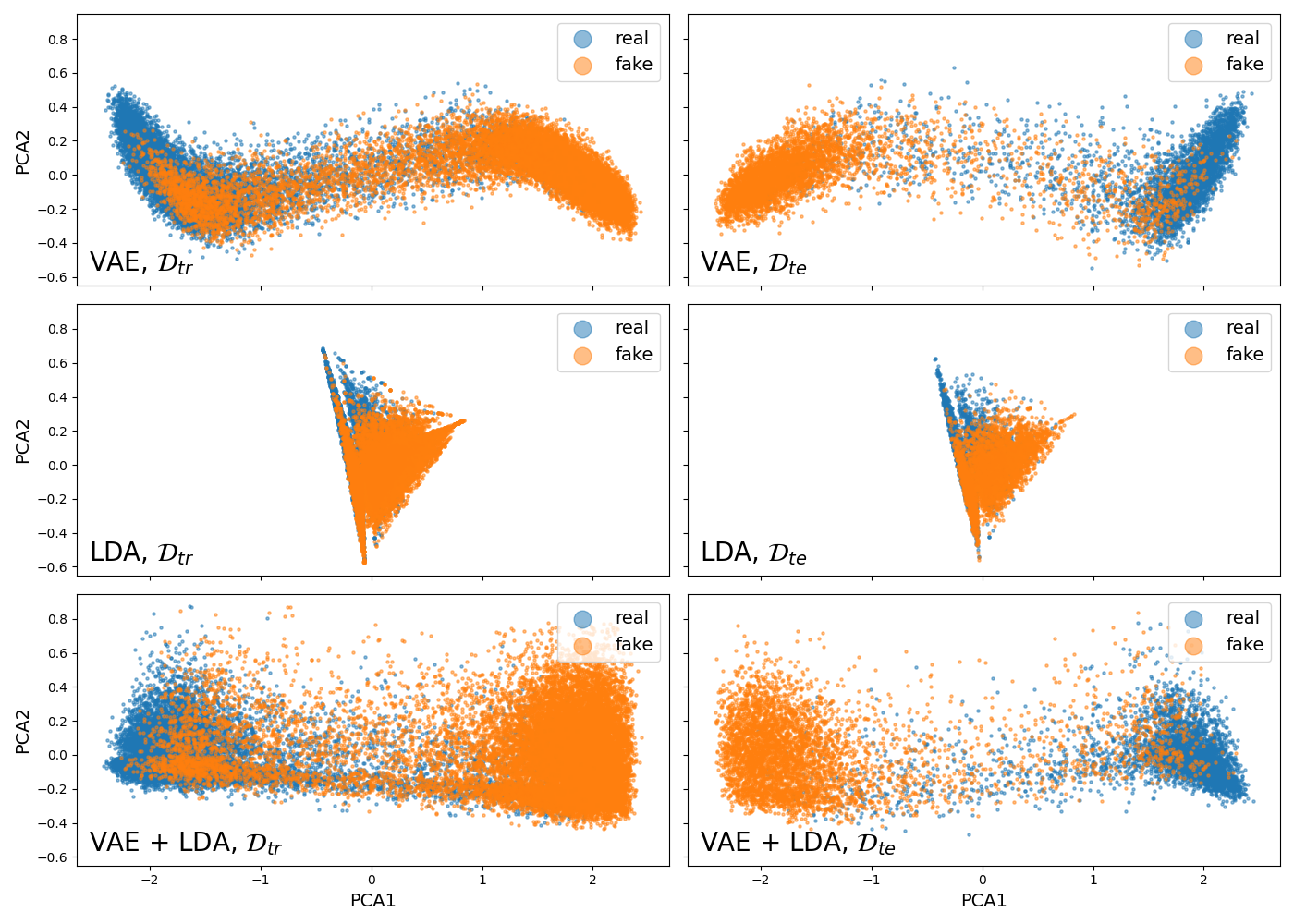}
\caption{\textbf{PCA Plot for ISOT dataset.} The first and second columns illustrate the application of PCA on training and test data. 
The first two rows correspond to the feature sets obtained by VAE and LDA separately, and the third row corresponds to their concatenation. }
\label{figs:pcaallisot}
\end{figure}

\begin{figure}[h]
\centering 
\includegraphics[trim=10 10 0 10,clip,width=0.50\textwidth]{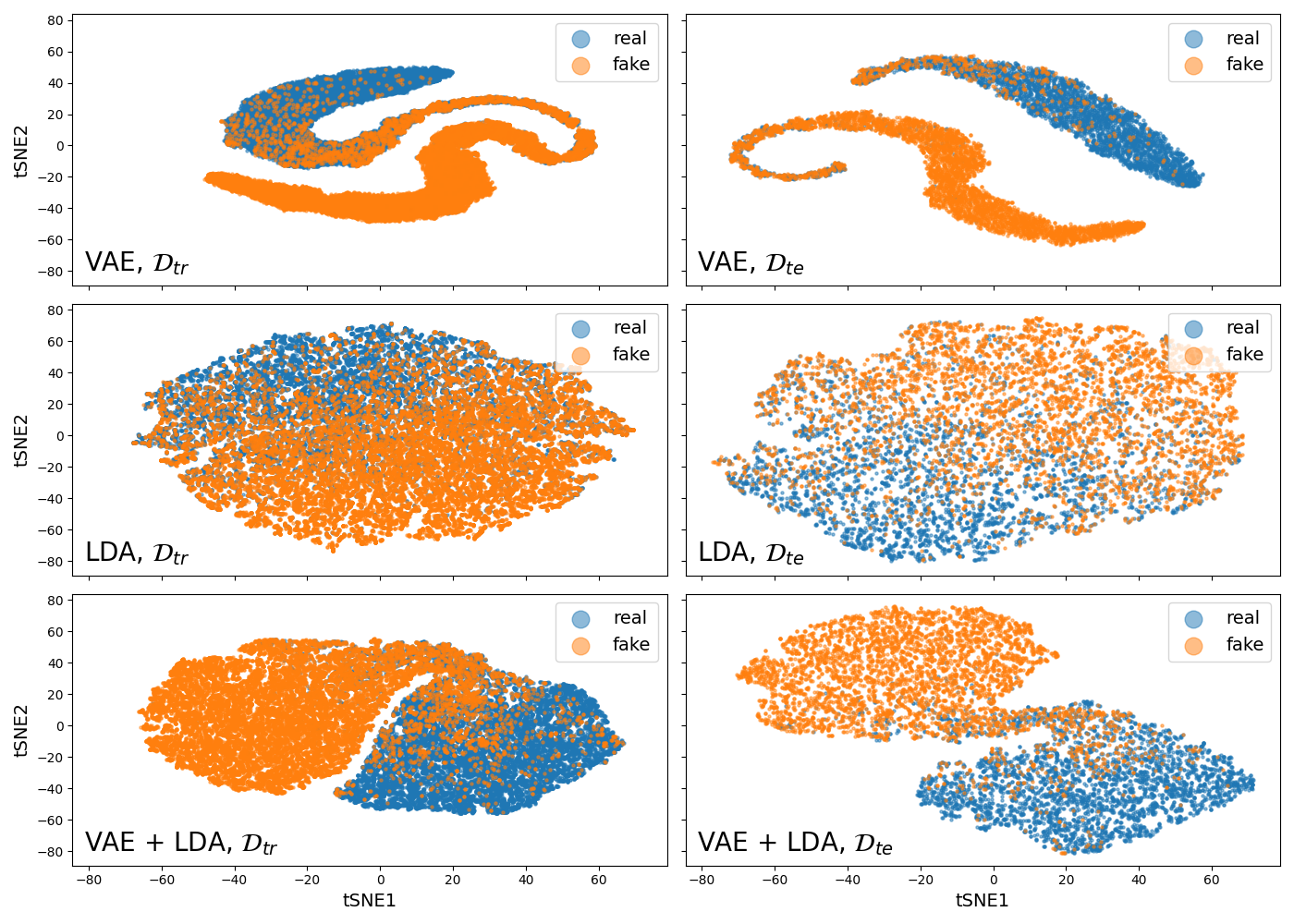}
\caption{\textbf{t-SNE Plot for ISOT dataset}. The first and second columns illustrate the application of t-SNE on training and test data. The first two rows correspond to the feature sets obtained by VAE and LDA separately, the third row shows the result on their concatenation. }
\label{figs:tsneISOT}
\end{figure}

% In this section we provide the application of dimensionality reduction methods on obtained feature sets and their concatenation.
Lastly, we qualitatively evaluated the separability of LDA, VAE, and combined LDA and VAE features in 2-dimensional embeddings for the ISOT dataset; the plots of other datasets show similar behavior.
% and can be found in the GitHub repository along with some technical and implementation details.
% github repository\footnote{\href{https://github.com/Marjan-Hosseini/Explainable-FakeNews-Detector}{https://github.com/Marjan-Hosseini/Explainable-FakeNews-Detector}}.
% \begin{figure}[h]
% \centering 
% \includegraphics[trim=10 10 0 10,clip,width=0.50\textwidth]{Figs/PCA_all_ISOT_n_topis_10_latent_dim_32_w2v_32_all_new.png}
% \caption{\textbf{PCA Plot for ISOT dataset.} The first and second columns illustrate the application of PCA on training and test data. 
% The first two rows correspond to the feature sets obtained by VAE and LDA separately, the third row shows the result on their concatenation. }
% \label{figs:pcaallisot}
% \end{figure}
We applied PCA and t-SNE to the features learned from the different architecture configurations with $K = 10$ and $w = n_f = 32$ (Figs.~\ref{figs:pcaallisot} and \ref{figs:tsneISOT}). 
% The feature sets obtained from VAE are drawn from a Gaussian, and the $\bm{\varphi}$ variables in LDA are Dirichlet distribution. 
We observe that in both PCA and t-SNE plots, the real and fake news posts are well separated by the combined VAE and LDA features. % i'm not totally convinced by this. do the other pca and tsne plots look better?

%Since PCA is a linear transformation and cannot capture the non-linear relationship between features 
% We provided the tSNE transformation for the same feature sets in Fig.~\ref{figs:tsneISOT} too. 
% For Covid dataset, we only plotted tSNE~\ref{figs:tsneCovid} because it visualizes the data better. 
% \begin{figure}[h]
% \centering 
% \includegraphics[trim=10 10 0 10,clip,width=0.50\textwidth]{Figs/tSNE_all_Covid_paper_tSNE.png}
% \caption{tSNE Plot for Covid dataset, the first and second columns illustrate the application of tSNE on training and test data. The first two rows correspond to the feature sets obtained by VAE and LDA separately, the third row shows the result on their concatenation. In all the plots $K = 10$ and $w = n_f = 32$.}
% \label{figs:tsneCovid}
% \end{figure}

% \hl{Another figure:}

% \section{Discussion}
% \label{sec:Discussion}

% \hl{discuss the potential scalabibility to other scenarios;
% discuss the integration with multi-modality datasets}

% In this section, some advantages and disadvantages of this work are discussed.

% \hl{Talk about future work on combining supervised LDA with deep models.} 

% \break

\section{Conclusion and Future Work}
\label{sec:concandfw}

We proposed \texttt{LDAVAE}, a combined LDA, supervised Bi-LSTM VAE, and classifier architecture for classifying fake news text. 
We use a Bayesian admixture model for topic modelling to increase the interpretability of the method, add informative features, and remove the necessity of costly manual topic selection. 
We justified our architecture with an extensive ablation study and evaluated performance of \texttt{LDAVAE} by comparing with $4$ SOTA baseline methods across $3$ datasets and showed highly competitive performance. 
%classifying individual feature sets as baselines and their concatenation and demonstrated that our model successfully improves the accuracy metrics in most classifiers. 
We then provided mechanisms to evaluate the interpretability of model parameters and class discrimination allowing for exploration of the model and features.
% In another experiment, we performed linear and non-linear dimensionality reduction methods on the obtained features and showed that our model could facilitate clustering in unsupervised settings.
%One aim of this work is increasing interpretability in the fake news detection framework, which we achieve by adding LDA to our model.
%Moreover, even if we use only the textual part of the news, our model improves performance metrics criteria together with most of the classifiers over the base models and outperforms or is comparable with SOTA.
% We have shown that the
% concatenation of obtained features, can separate these two classes more clearly.
% , so we expect clustering methods can be applied more efficiently.

The main disadvantage of our modelling assumptions is that we cannot extract features from multimedia data.
In addition, training a separate probabilistic model requires additional computation, however understanding the model and important features are crucial for providing explanations and establishing user trust.
% , and users can decide if they are willing to sacrifice a few percentage in accuracy.
% is often considered more important 
% One possible improvement can be adopting extra text preprocessing, especially to handle noisy datasets such as Twitter. 
% One possible way to improve the results is by adopting extra text preprocessing, especially to handle noisy datasets such as Twitter. 
More sophisticated natural language architectures that use self-attention, like transformers, should provide a mechanism for improving accuracy.
%For improving the black box accuracy in the future, 
% opportunity for future is to use more sophisticated embedding models like BERT~\cite{devlin2018bert} instead of word2vec. 
%we suggest BERT which uses transformer layers with self-attention mechanisms
% ~\cite{vaswani2017attention}
%that could enhance the efficiency in datasets with shorter maximum lengths.
Integrating transformers, which typically model text as an ordered sequence, with topics that are intrinsically unordered is a priming area of future work~\citep{wang2020friendly}.
% We can incorporate more discriminative textual-based features to increase the accuracy or alter deep architecture~\cite{mozhdehi2018deep}.
Also, alternatives to VAEs, like architectures based on normalizing flows, should be investigated in the context of fake news detection~\citep{bohm2020probabilistic}. 
For example, the Probabilistic Auto-Encoder leverages normalizing flows to achieve better reconstruction than traditional VAEs while retaining high sample quality. 
% A higher quality of latent space would potentially improves the features obtained by VAE.
%normalizing flow methods can transform the latent space to alleviate the VAE performance by  improving the reconstructure error
%and feature selection techniques suitable for high dimensional data can increase the accuracy and interpretability~\cite{bohm2020probabilistic}.
Future work may also consider applying feature selection techniques suitable for high dimensional data or selecting the dimension of the autoencoder latent space~\citep{brankovic2018distributed, hosseini2018feature}, and using the text provided in reviews~\citep{qian2018neural} and conversation graphs~ \citep{10.1007/978-3-030-74296-6_8} of live events participants~\citep{brambillaDataDIB, JAVADIANSABET2021100140} to detect the disinformation propagation on inner circles~\citep{de2020systematic}.

% \vspace{-0.8cm}
% In the future, we plan to investigate the conversation graphs~\cite{10.1007/978-3-030-74296-6_8} of the live events participants~\cite{JAVADIANSABET2021100140} to detect the disinformation propagation on inner circles~\cite{de2020systematic}.

% \section*{Declaration of Competing Interest}

% The authors declare that there is no conflict of interest in all aspects of this manuscript preparation and data analysis.

% \newpage

%% Loading bibliography style file
%\bibliographystyle{model1-num-names}
% \bibliographystyle{cas-model2-names}

% Loading bibliography database
% \bibliography{cas-refs}

% \vspace{-0.5cm}

% \begin{figure*}[]
% \centering 
% \includegraphics[width=1\textwidth]{Figs/radar_all_with_covid_examples.png}
% \caption{\textbf{Interpreting $\bm{\varphi}$.} Averaged posterior means of $\varphi_i$ shows differences in real and fake news post topic usage across the ISOT (left), Covid (middle), and Twitter (right) datasets ......}
% \label{figs:radar_with_cov_ex}
% \end{figure*}

\section*{Declaration of Competing Interest}

The authors declare that there is no conflict of interest in all aspects of this manuscript preparation and data analysis.

\printcredits

% \pagebreak
% \newpage

%% Loading bibliography style file
\bibliographystyle{model1-num-names}
%\bibliographystyle{model2-num-names}
% \bibliographystyle{unsrt}
% \bibliographystyle{apalike}
% \bibliographystyle{elsarticle-harv}
% \bibliographystyle{cas-model2-names}

% \bibliographystyle{model5-names}
% \biboptions{authoryear}

% \bibliographystyle{cas-model2-names}

% Loading bibliography database
% \bibliography{cas-refs}
\bibliography{References.bib}

%\vskip3pt

\pagebreak

\bio{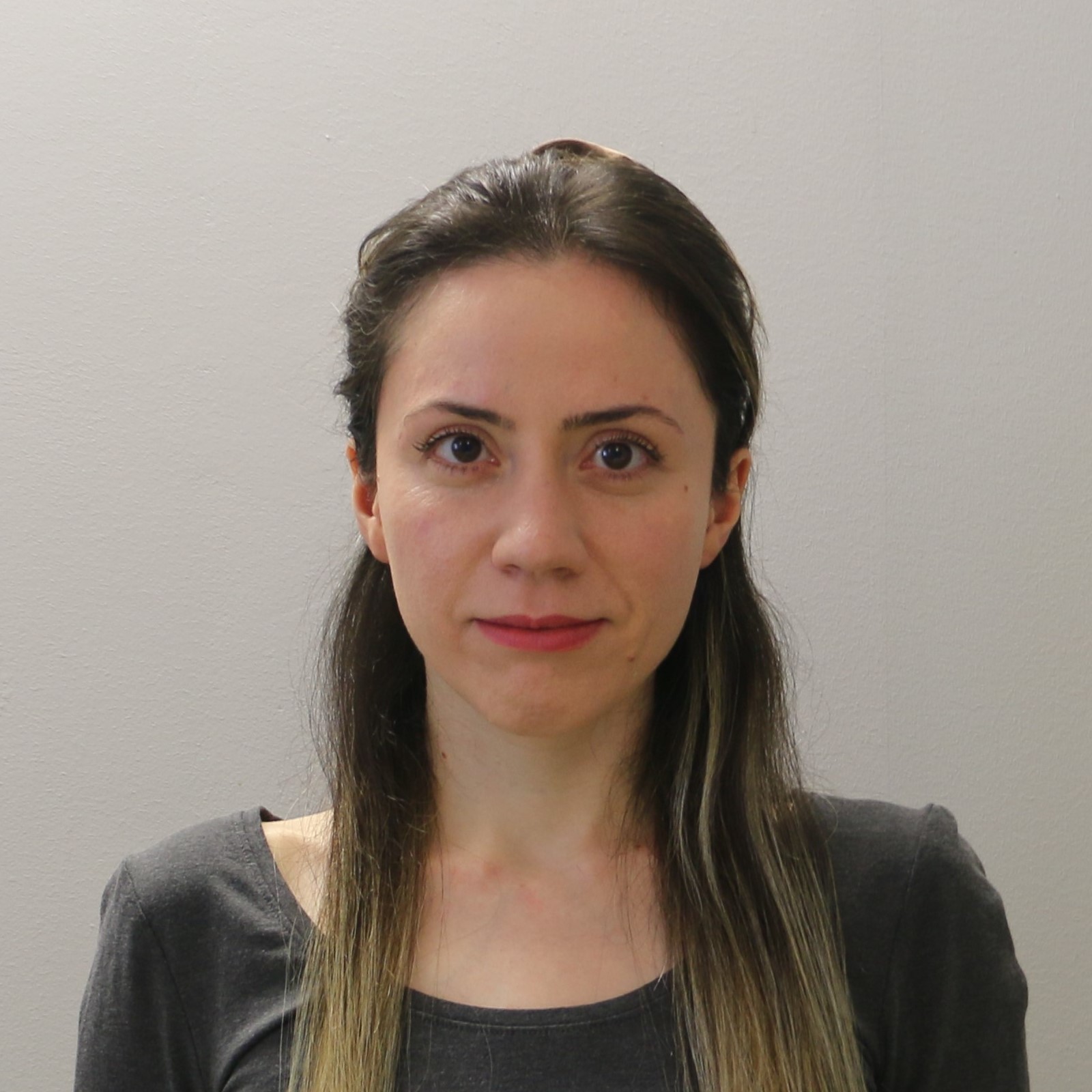}
\textbf{Marjan Hosseini} received her baccalaureate in computer engineering from Shomal University, Iran, and her master's degree in computer science and engineering from Politecnico di Milano (POLIMI) in 2018. Afterwards, she joined DEEP-SE Lab at POLIMI and worked in natural language processing and machine learning. Currently, she is working towards her PhD at the University of Connecticut in the Computer Science and Engineering Department. Her research interests lie in Bayesian and probabilistic methods in computational biology, machine learning and bioinformatics.
\endbio

\bio{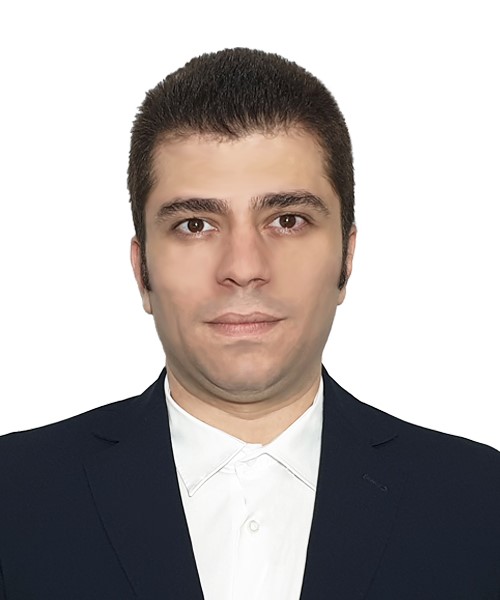}
\textbf{Alireza Javadian Sabet} is a PhD student in the Department of Informatics and Networked Systems at the University of Pittsburgh. He received his master's degree in computer science and engineering from POLIMI.
He worked as a research fellow at the Data Science Lab and DEpendable Evolvable Pervasive Software Engineering (DEEP-SE) group at POLIMI. 
His research interests lie in computational social science, knowledge extraction from social networks, business analysis, and context-aware recommender systems.
% received his master's in Computer Science and Engineering from Politecnico di Milano in 2019. He worked on analyzing social media as his master's thesis and joined Data Science Lab at Politecnico di Milano (POLIMI) afterwards to continue his research. In 2019, he also joined DEpendable Evolvable Pervasive Software Engineering (DEEP-SE) group at POLIMI as a research fellow, and has been studying transportation systems in Europe. His research interests lie in Knowledge Extraction from Social Networks, Business Analysis and Context-aware Recommender Systems.

\endbio

\bio{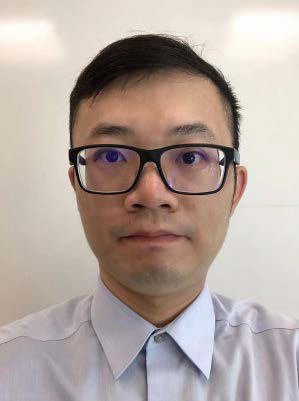}
\textbf{Suining He} is currently working as an assistant professor at the Department of Computer Science and Engineering, University of Connecticut. He received his PhD degree in computer science and engineering at the Hong Kong University of Science and Technology, and worked as a postdoctoral research fellow at the Real-Time Computing Lab (RTCL) at the University of Michigan, Ann Arbor. His research interests include ubiquitous and mobile computing,
crowdsourcing, and big data analytics.

\endbio

\bio{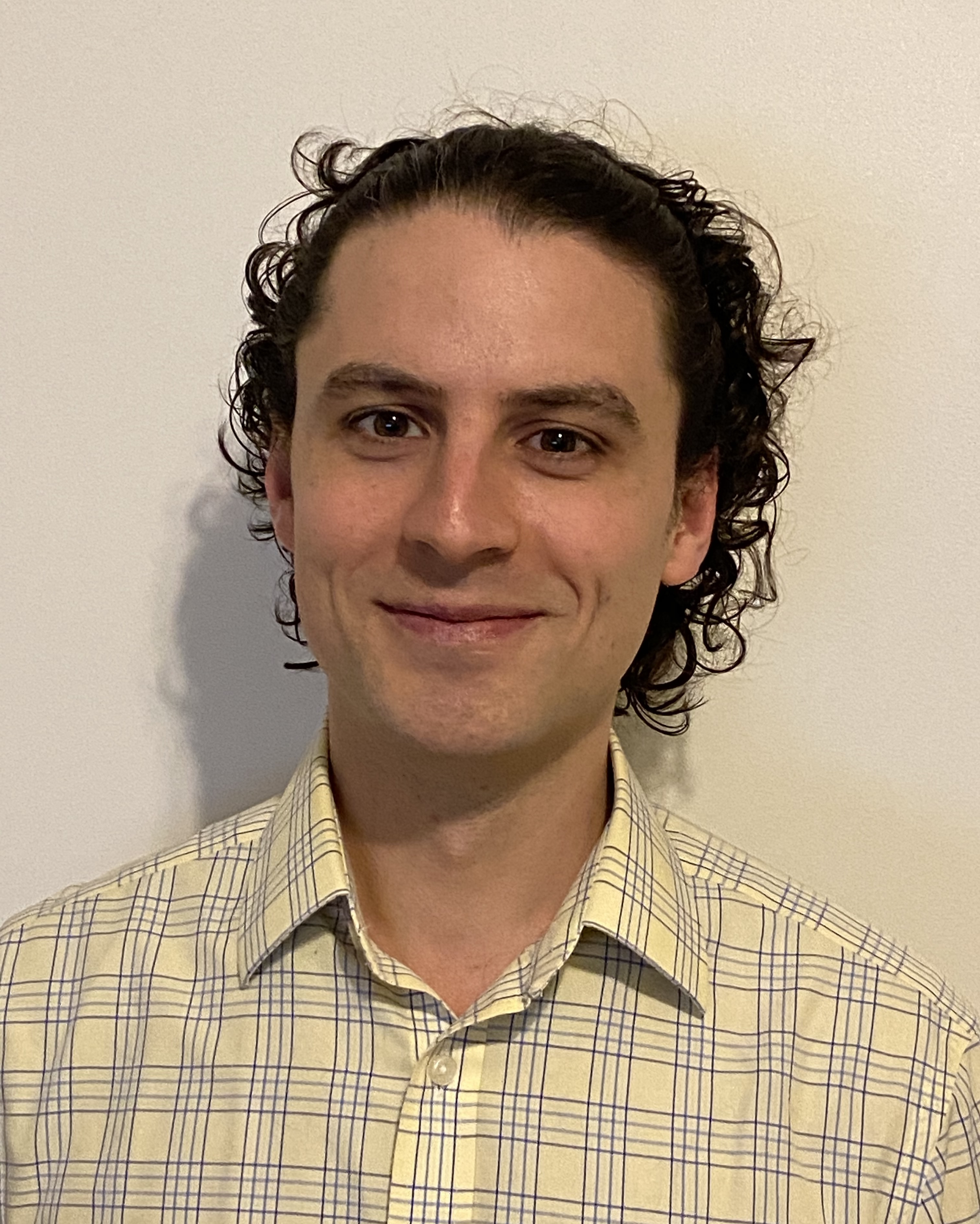}
\textbf{Derek Aguiar} is an assistant professor in the Computer Science and Engineering Department at the University of Connecticut. He graduated from the University of Rhode Island with baccalaureates in computer engineering and computer science, received his PhD in computer science from Brown University, and completed his postdoctoral work at Princeton University. His research aims to develop Bayesian machine learning models, scalable inference methods, and graph theoretic algorithms to better understand high-dimensional data, particularly in the areas of genomics and genetics.

\endbio

\end{document}